\documentclass[a4paper,fleqn]{cas-dc}
\usepackage{amssymb}
\usepackage{hyperref}
\usepackage{graphicx}
\usepackage[caption=false]{subfig}
\usepackage{lipsum}
\usepackage{lscape}
\usepackage{amsmath}
\usepackage{multirow}
\usepackage[figuresright]{rotating}
\usepackage{lineno}
\usepackage{algorithm} 
\usepackage{algpseudocode} 
\usepackage{comment}
\usepackage{lineno}
\usepackage{nomencl}
\makenomenclature
\usepackage{calc}
\usepackage[T1]{fontenc}
\usepackage{relsize}

\usepackage{natbib}
\setcitestyle{authoryear,open={(},close={)},citesep={;}} 

\hypersetup{
  colorlinks,
  citecolor=Violet,
  linkcolor=Red,
  urlcolor=Blue}

\newcolumntype{P}[1]{>{\centering\arraybackslash}p{#1}}

\def\tsc#1{\csdef{#1}{\textsc{\lowercase{#1}}\xspace}}
\tsc{WGM}
\tsc{QE}
\tsc{EP}
\tsc{PMS}
\tsc{BEC}
\tsc{DE}

\begin{document}
\let\WriteBookmarks\relax
\def\floatpagepagefraction{1}
\def\textpagefraction{.001}
\shorttitle{A Systematic Review on Label-Efficient Learning in Agriculture}
\shortauthors{Li et~al.}

\title [mode = title]{Label-Efficient Learning in Agriculture: A Comprehensive Review}

\author[1]{Jiajia Li}\ead{lijiajia@msu.edu}
\author[1]{Dong Chen}\ead{chendon9@msu.edu}
\author[1]{Xinda Qi}\ead{qixinda@msu.edu}
\author[2]{Zhaojian Li}\ead{lizhaoj1@egr.msu.edu}
\author[3]{Yanbo Huang}\ead{yanbo.huang@usda.gov}
\author[4]{Daniel Morris}\ead{dmorris@msu.edu}
\author[1]{Xiaobo Tan}\ead{xbtan@egr.msu.edu}

\address[1]{Department of Electrical and Computer Engineering, Michigan State University, East Lansing, MI 48824, USA}
\address[2]{Department of Mechanical Engineering, Michigan State University, East Lansing, MI 48824, USA}
\address[3]{United States Department of Agriculture Agricultural Research Service, Genetics and Sustainable Agriculture Research Unit, Mississippi State, MS 39762, USA}
\address[4]{Department of Biosystems and Agricultural Engineering
, Michigan State University, East Lansing, MI 48824, USA}

\address{* Zhaojian Li (\textit{lizhaoj1@egr.msu.edu}) is the corresponding author.}

\begin{abstract}
The past decade has witnessed many great successes of machine learning (ML) and deep learning (DL) applications in agricultural systems, including weed control, plant disease diagnosis, agricultural robotics, and precision livestock management. Despite tremendous progresses, one downside of such ML/DL models is that they generally rely on large-scale labeled datasets for training, and the performance of such models is strongly influenced by the size and quality of available labeled data samples. In addition, collecting, processing, and labeling such large-scale datasets is extremely costly and time-consuming, partially due to the rising cost in human labor. Therefore, developing label-efficient ML/DL methods for agricultural applications has received significant interests among researchers and practitioners. In fact, there are more than 50 papers on developing and applying deep-learning-based label-efficient techniques to address various agricultural problems since 2016, which motivates the authors to provide a timely and comprehensive review of recent label-efficient ML/DL methods in agricultural applications. To this end, we first develop a principled taxonomy to organize these methods according to the degree of supervision, including weak supervision  (i.e., active learning and semi-/weakly- supervised learning), and no supervision (i.e., un-/self- supervised learning),  supplemented by representative state-of-the-art label-efficient ML/DL methods. In addition,  a systematic review of various agricultural applications exploiting these label-efficient algorithms, such as precision agriculture, plant phenotyping, and postharvest quality assessment,  is presented. Finally, we discuss the current problems and challenges, as well as future research directions. A well-classified paper list that will be actively updated can be accessed at \url{https://github.com/DongChen06/Label-efficient-in-Agriculture}.
\end{abstract}


\maketitle

\section{Introduction}
\label{sec:intro}
Smart farming (also referred to as smart agriculture) \citep{walter2017smart, wolfert2017big, moysiadis2021smart} integrated with a range of recent information and communication technologies (ICT), including unmanned aerial/ground vehicles (UAVs/UGVs), image processing, machine learning, big data, cloud computing, and wireless sensor networks (WSNs), has emerged as a promising solution to boosting the agricultural outputs, increasing farming efficiency as well as the quality of the final product \citep{walter2017smart, wolfert2017big}. With smart farming, farmers can make informed planting, tending and harvesting decisions with data collected from smart sensors and devices. However, extracting relevant and useful information from diverse data sources and especially imaging data, is challenging.  Traditional data mining techniques are often unable to reveal meaningful insights from these complex data \citep{wolfert2017big}. 

Deep learning (DL \citep{lecun2015deep}), on the other hand, has shown great capabilities in processing complex, high-dimensional data with numerous successful applications \citep{he2016deep, he2017mask, dosovitskiy2020image}. 
In particular, DL methods have demonstrated remarkable feature extraction and pattern classification capabilities,  learning high-quality image representations and achieving promising performance in various agricultural applications \citep{kamilaris2018deep}, including weed control \citep{chen2022performance}, plant disease detection \citep{xu2021style, fan2022leaf}, postharvest quality assessment \citep{zhou2022deep}, and robotic fruit harvesting \citep{chu2021deep,zhang2022algorithm, chu2023o2rnet}, among others.
In spite of the promising progress, the aforementioned approaches are mainly based on supervised training that is universally acknowledged as data-hungry, and the performance of such supervised methods is highly dependent on large scale and high quality labeled datasets \citep{sun2017revisiting}. For example, in computer vision tasks such as object detection and semantic segmentation, the models are generally pre-trained on large-scale image datasets in a supervised fashion with large volumes of labeled images, such as ImageNet \citep{deng2009imagenet}, Microsoft COCO \citep{lin2014microsoft}, and PlantCLEF2022 \citep{goeau2022overview}. However, the collection and annotation of such datasets are extremely time-consuming, resource-intensive, and expensive.  It is highly desirable to avoid repeating this for new applications in farming.   

To mitigate the costly and tedious process in data annotation, there has been an emerging ML field that focuses on developing weak supervision \citep{zhou2018brief, van2020survey} or even no supervision \citep{jing2020self} approaches to learn feature representation from large-scale \emph{unlabeled} data. Specifically, in weak supervision, a set of unlabeled data samples or data samples with coarse labels, which are cheap and easier to obtain,  along with small portions of labeled samples are jointly employed to train the ML/DL models \citep{zhou2018brief, van2020survey}. In unsupervised approaches, the models are trained with large-scale unlabeled data without requiring any human-annotated labels \citep{min2018survey, jing2020self}. The goal of this review paper is thus to survey such label-efficient learning methods \citep{shen2022survey}, along with their applications to agricultural systems, with a focus on DL techniques.

To date, several surveys of label-efficient learning approaches have been published \citep{zhou2018brief, van2020survey, min2018survey, jing2020self}.  In \cite{zhou2018brief}, the authors reviewed some advancements of weakly supervised learning, in which three types of approaches with weak supervision were introduced: incomplete supervision, inexact supervision, and accurate supervision. In \cite{jing2020self}, the authors focused on self-supervised learning methods for general visual feature learning (i.e., inputs to networks are images or videos). However, these surveys introduce approaches in a relatively isolated way.  There are also a few survey papers that focus on a specific aspect/task of these label-efficient approaches  \citep{schmarje2021survey, shen2022survey}. For instance, a review of semi-, self-, and unsupervised learning for image classification tasks was presented in \citep{schmarje2021survey}, in which 34 common methods are implemented and compared. In \cite{shen2022survey},  semi-supervised and weakly-supervised learning approaches for image segmentation tasks, including semantic segmentation, instance segmentation, and panoptic segmentation, were reviewed. Therefore, a review on label-efficient learning with comprehensive coverage of the methodology and the corresponding tasks is still lacking.

Furthermore, despite the rapid growth in smart agriculture, review papers on label-efficient learning for agricultural applications have been scarce. In a conference paper \citep{fatima2021semi}, the authors reviewed recent advanced semi-supervised learning algorithms (from 6 conference papers and 6 journal papers) for smart agriculture. However, they only reviewed the semi-supervised approaches, while the recent advances in other label-efficient learning areas, such as weakly- and unsupervised learning, are lacking. Recently, \cite{yan2022unsupervised} reviewed the unsupervised and semi-supervised learning approaches for plant system biology. Compared to the aforementioned review papers, our work differs in the following aspects. Firstly, \cite{yan2022unsupervised} mainly focused on the applications of plant system biology, while we present a broader range of agricultural applications, including precision agriculture, plant phenotyping, and postharvest quality assessment. Secondly, in \citep{yan2022unsupervised}, the surveyed papers were mostly related to analyzing plant omics data (i.e., genome, metabolome, phenome, proteome, and transcriptome), so the techniques of self-supervised learning and semi-supervised learning approaches were mostly conventional machine learning approaches developed for low-dimensional data, while we focused more on image feature learning (i.e., inputs are high-dimensional RGB images) based on advanced deep neural networks \citep{simonyan2014very, he2016deep}. Lastly, we develop a principled taxonomy to organize these methods according to the degree of supervision, including weak supervision (i.e., active and semi-/weakly- supervised learning), and no supervision (i.e., un/self- supervised learning). Given the rapid development in label-efficient learning, our work attempts to comprehensively review the state-of-the-art algorithms with focuses on agricultural applications, covering the most prominent and  relevant works in a principled fashion.

 In this survey, we first propose a new taxonomy of label-efficient algorithms to organize different conceptual and methodological approaches according to the degree of required supervision. In addition, we summarize the most representative methods along with the developed and publicly available packages/tools. Furthermore, we review recent advances in weak and no supervision learning and their applications in agriculture, including precision agriculture, plant phenotyping, and postharvest management. Last but not least, we discuss the remaining challenges and potential future directions. This review will be beneficial for researchers who are new to this field as well as for those who have a solid understanding of the main approaches but want to work on innovative applications in the agricultural space.



\section{Methods}
\label{sec:Methods}


\begin{figure*}[!ht]
  \centering
\includegraphics[width=0.80\textwidth]{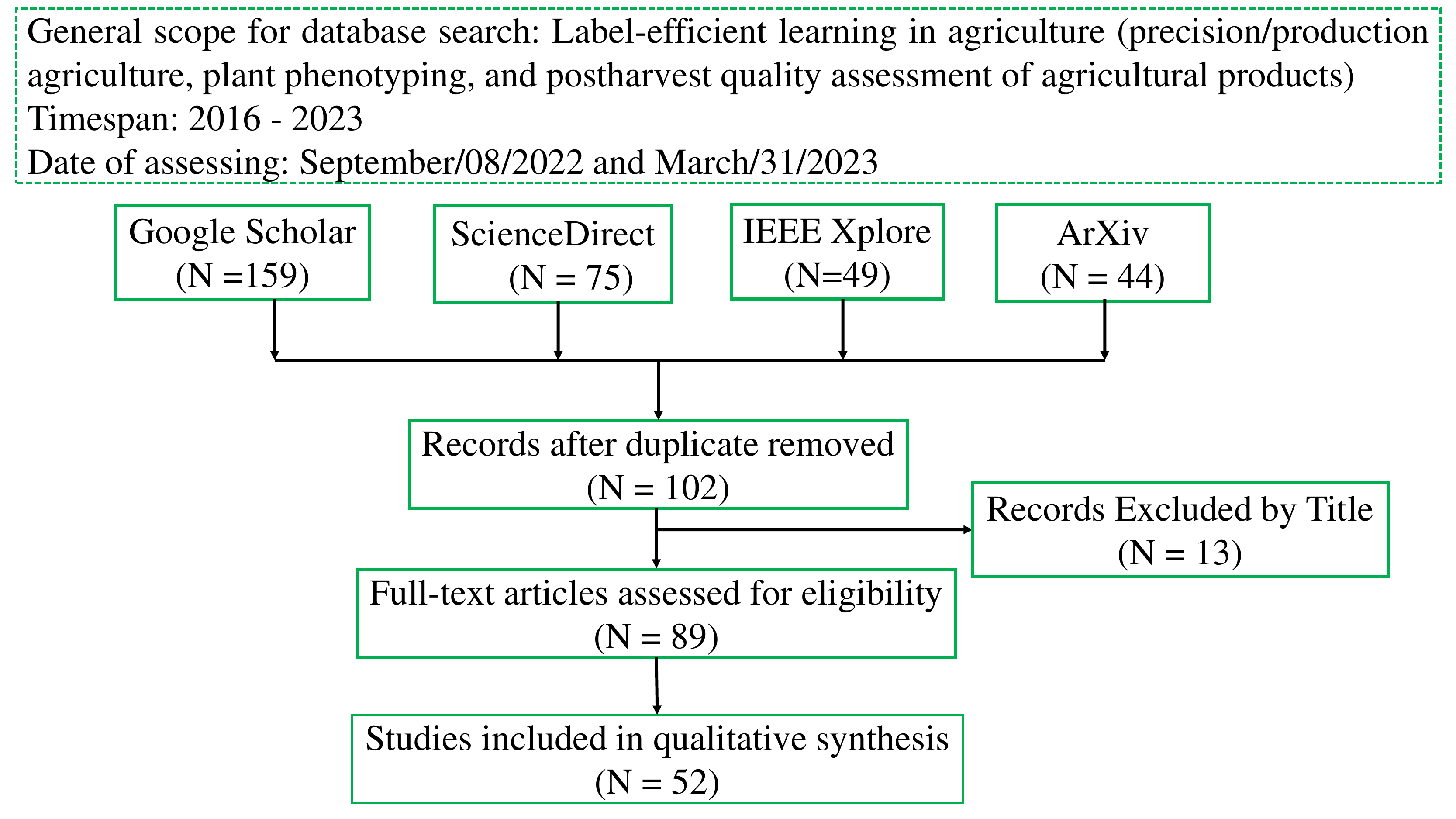}
  \caption{The PRISMA guideline flowchart used in this review. The figure first row illustrates initially selected articles based on the keywords that enhanced the initial
filtering before other exclusion criteria are applied.}
  \label{fig:PRISMA}
  \vspace{-10pt}
\end{figure*}

In this survey, the conventional PRISMA (\underline{P}referred \underline{R}eporting \underline{I}tems for \underline{S}ystematic \underline{R}eviews and \underline{M}eta-\underline{A}nalysis) method \citep{moher2009preferred} is used to thoroughly and systematically collect related literature, by exploiting the recommended methods for literature collection, as well as the inclusion and exclusion criteria \citep{moher2009preferred, lu2022generative}.
Specifically, the databases for the literature collection are chosen to ensure the comprehensiveness of the review. Firstly, major scientific databases (e.g., Web of Science, ScienceDirect, Springer, and Elsevier) are selected for searching related topics. Secondly, various mainstream scientific article search engines and databases, including Google Scholar, IEEE Xplore, and the open-access paper platform ArXiv\footnote{Arxiv: \url{https://arxiv.org/}}, are utilized to expand the search coverage and collect more recent literature, which is crucial for identifying emerging label-efficient ML/DL approaches and their agricultural applications.

Once the databases for the literature collection are determined, an \textit{inclusion} criterion is applied to the article search in the identified databases and search engines. Specifically, keyword search is first conducted for a preliminary article collection, using a combination of two groups of words as keywords. The first group of words is selected in the label-efficient learning field, such as ``active learning'', ``semi-supervised learning'', ``weakly-supervised learning'', ``self-supervised learning'', ``unsupervised learning'', ``label-free'', and ``label-efficient learning''. The second group of words is selected in the agricultural field, such as ``agricultural applications'', ``precision agriculture'', ``weed'', ``fruit'', ``aquaculture'', ``plant phenotyping'' and ``postharvest quality assessment''. Keyword operators, such as ``AND'' and ``OR'' are also used in the  process to improve the efficiency and diversity of the keyword search. Furthermore, references and citations of the selected articles from the keyword search are also included to expand the initial \textit{inclusion} stage. 

To select the most relevant literature for the review paper, several \textit{exclusion} criteria are then used to filter the articles obtained from the preliminary collection. The first exclusion criterion is the publication date. We mainly focus on articles published in the last eight years (2016-2023) because label-efficient learning based on deep learning techniques in agriculture is relatively recent and related technologies are evolving rapidly. For older publications, we still include those with high citation indexes, considering their significant impact on others' work. The second exclusion criterion is the type of papers in the preliminary collection. We mainly focus on  research articles in highly ranked journals and filter out other types of papers, such as reports and meeting abstracts due to generally lower technical contribution and completeness. We also remove repeated literature resulting from searches in multiple databases and search engines.  After literature collection and screening, we finally obtain 52 research articles (see Fig.~\ref{fig:PRISMA} for the process) for label-efficient learning in the agriculture domain, which have been listed and will be actively updated on our GitHub repository: \url{https://github.com/DongChen06/Label-efficient-in-Agriculture}.

\begin{figure}[!ht]
  \centering
\includegraphics[width=0.45\textwidth]{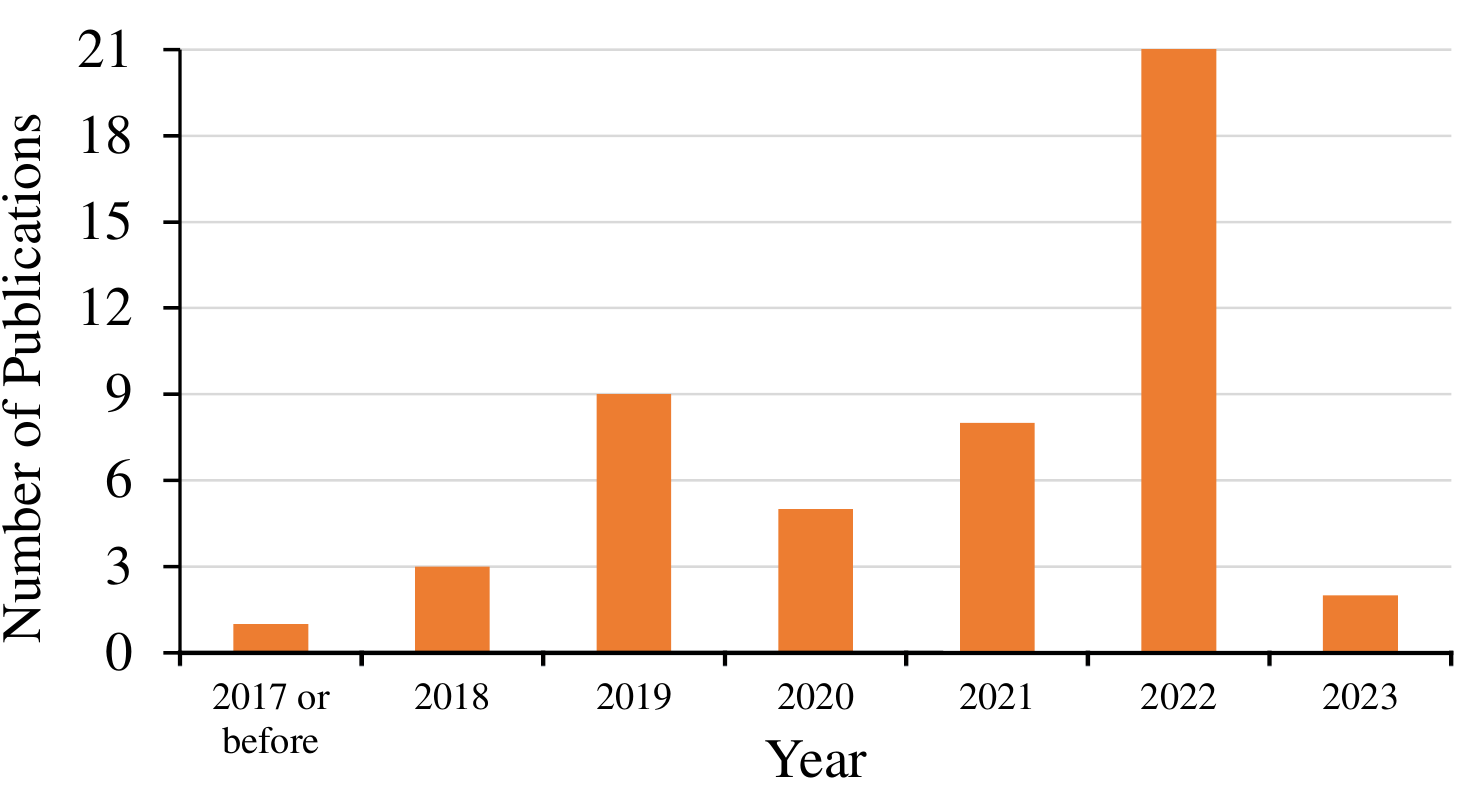}
  \caption{Numbers of publications over the years.}
  \label{fig:number_papers}
  \vspace{-10pt}
\end{figure}

The literature collections are then organized based on publication time to show the overall popularity trends of label-efficient learning in agriculture based on DL techniques. Fig.~\ref{fig:number_papers} shows the number of collected articles in different years within the time scope we focus on. It is observed that the topic of label-efficient learning has gained increasing attention in agricultural research from 2016 to the present (March 2023), demonstrating the significance and necessity of this review to cover the most prominent and relevant work.


\begin{figure*}[!ht]
  \centering
\includegraphics[width=0.95\textwidth]{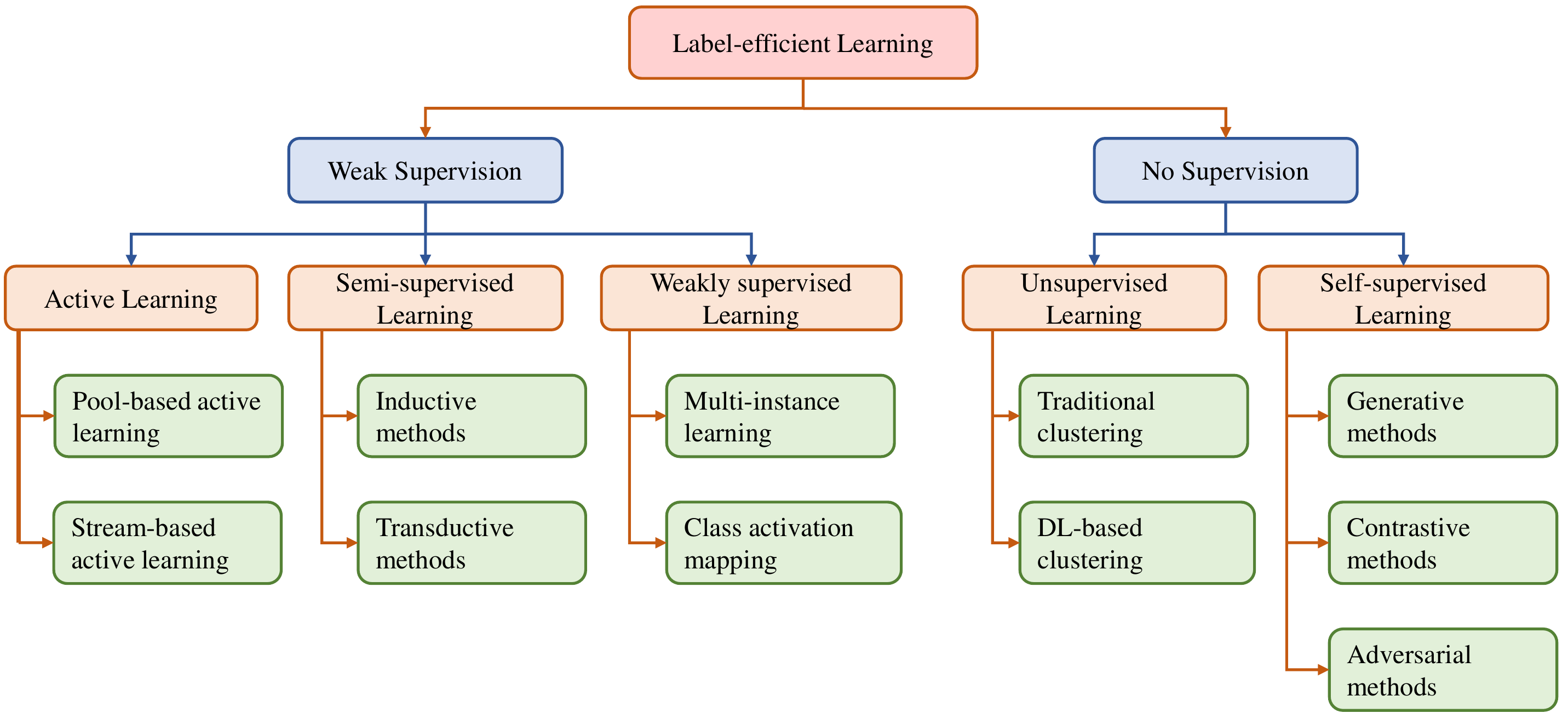}
  \caption{A comprehensive taxonomy of representative \textit{label-efficient learning} techniques. Our taxonomy encompasses two primary categories (shown in blue boxes), each consisting of multiple sub-areas (shown in green boxes).}
  \label{fig:taxonomy}
  \vspace{-10pt}
\end{figure*}

\section{Taxonomy of Label-Efficient Learning Methods}
\label{sec:concepts}
In this section, we will introduce the most representative label-efficient learning algorithms used in general computer vision (CV) tasks. As shown in Fig.~\ref{fig:taxonomy}, the proposed taxonomy encompasses two main categories: \emph{weak supervision} and \emph{no supervision}, each consisting of multiple sub-areas (green boxes in Fig.~\ref{fig:taxonomy}).

The training objective function for algorithms under weak supervision or no supervision can be represented as the following unified function: 
\begin{equation} \label{eqn:general_equation}
   \min_{\theta}\quad \lambda_l\cdot\sum_{(\mathbf{x},y) \in \mathcal{D}_L} \mathcal{L}_{sup} (\mathbf{x}, y, \theta) + \lambda_u\cdot \sum_{\mathbf{x} \in \mathcal{D}_U} \mathcal{L}_{unsup} (\mathbf{x}, \theta),
\end{equation}
where the ML/DL model weights $\theta$ is optimized.  $\mathcal{D}_L$ and $\mathcal{D}_U$ represent the (weakly) labeled and unlabeled datasets, respectively. The input data and the corresponding (weak) labels are represented by $\mathbf{x}$ and $y$. The weight coefficients for labeled and unlabeled data are denoted by $\lambda_l$ and $\lambda_u$, respectively. When there is no supervision, $\lambda_l$ is set to zero, indicating that only unlabeled data is used in the training process. However, in algorithms under weak supervision, both (weakly) labeled and unlabeled data sets are utilized to facilitate representation learning. For instance, in semi-supervised self-training \citep{lee2013pseudo} (Section~\ref{sec:semi}), ML/DL models are trained on joint samples with human-annotated labels and  pseudo-labels generated from unlabeled data.


\subsection{Weak Supervision}
Weak supervision refers to machine learning methods that utilize both labeled and unlabeled samples, where the labels may be incomplete or inaccurate or the unlabeled samples may be large in quantity  \citep{zhou2018weakly}.  These methods can be divided into three subcategories: active learning, semi-supervised learning, and weakly supervised learning. Specifically, active learning involves an iterative process of selecting the most informative data points for annotation to maximize model performance while minimizing the cost of human labeling. In semi-supervised learning, both labeled and unlabeled data are utilized for model training, with the goal of improving performance beyond what is achievable with only labeled data. Finally, weakly supervised learning involves training models with imperfect or incomplete labels, which can be easier and cheaper to obtain as compared fully annotated data. We next review more details about these methods.

\subsubsection{Active learning}\label{sec:al}
\begin{figure}[htp]
\subfloat[Stream-based active learning]{%
\includegraphics[width=0.45\textwidth]{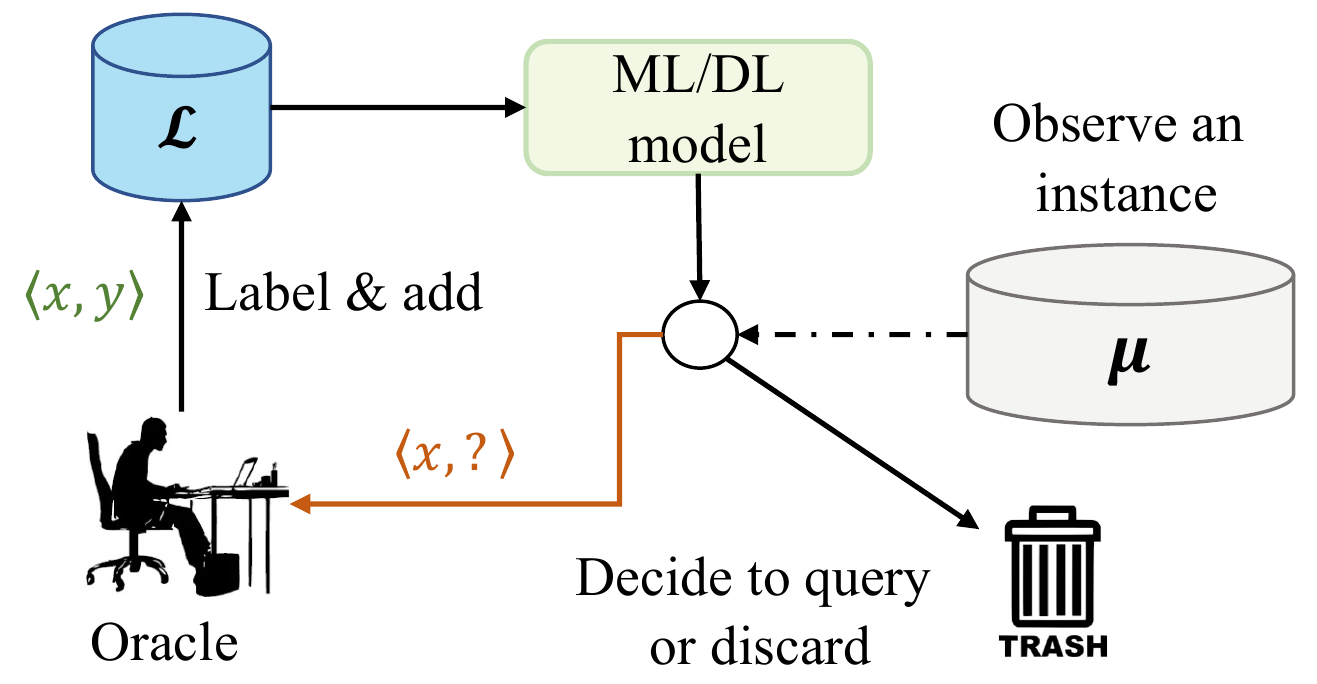}%
}

\subfloat[Pool-based active learning]{%
\includegraphics[width=0.45\textwidth]{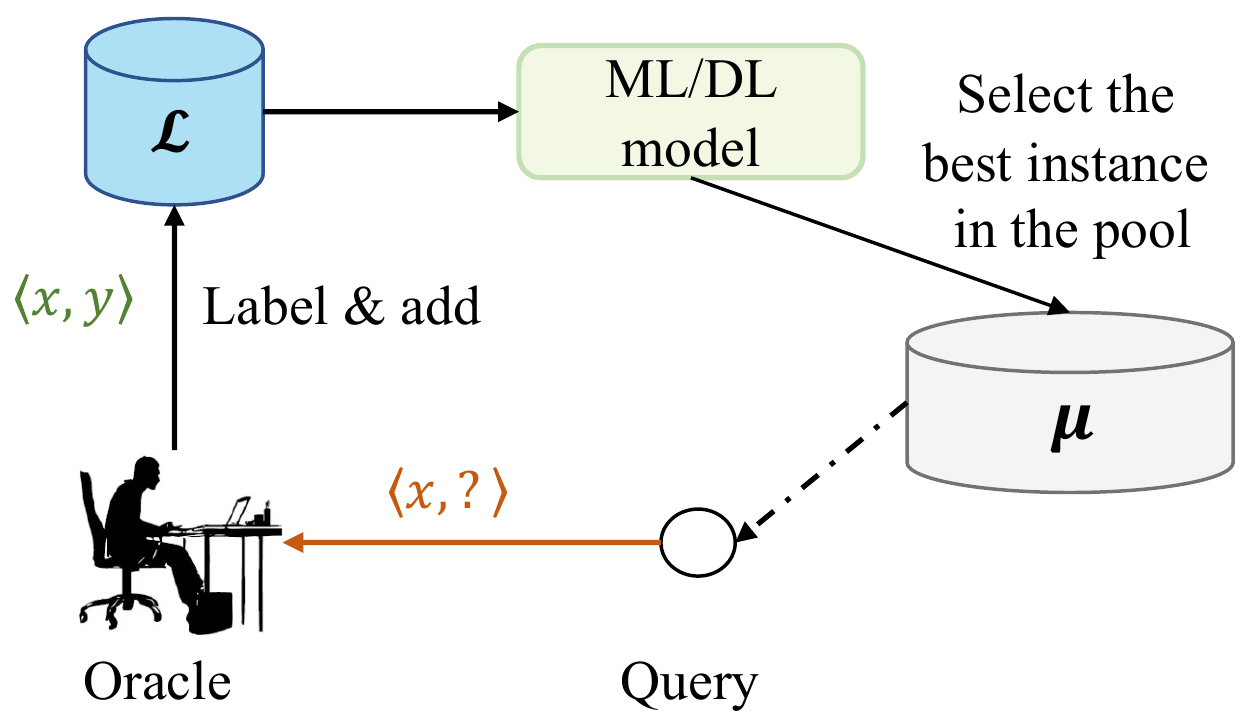}%
}
\caption{Diagrams of the active learning algorithms: stream-based selective sampling and pool-based sampling active learning. $\mathcal{L}$ and $\mu$ represent the labeled and unlabeled datasets, respectively.}
\label{fig:al}
\end{figure}

Active learning \citep{settles2009active, ren2021survey} aims to achieve maximum performance gains with minimum annotation effort. With a large pool of unlabeled samples, active learning selects the most informative samples and then requests labels from an ``oracle'' (typically a human annotator) to minimize the labeling cost. As shown in Fig.~\ref{fig:al}, active learning can be further categorized into two types: stream-based active learning and pool-based active learning. 

In the stream-based methods (Fig.~\ref{fig:al} (a)), one instance (i.e., sample) is selected at a time for the query sequentially from the input data source, and the ML/DL model needs to make a decision whether to query or discard it individually. This approach is extremely useful for resource-intensive scenarios, such as training and inference on mobile and embedded devices. On the other hand, pool-based active learning (Fig.~\ref{fig:al} (b)) ranks and selects the best query from the entire unlabeled set \citep{settles2009active}, which is exploited in most real-world applications where large amounts of unlabeled samples can be accessible and processed at once. Unless specifically noted, the subsequent discussions will focus on the pool-based methods. 

In pool-based active learning, $D_L = \{X, Y \}$ is defined as the labeled dataset with $m$ samples, where $\text{x} \in X$ and $\text{y} \in Y$ represent the samples and their labels, respectively. $D_U = \{\mathcal{X}, \mathcal{Y} \}$ is the unlabeled dataset with $n$ samples, where $x \in \mathcal{X}$ and $y \in \mathcal{Y}$ denote the sample space and label space. In active learning settings ($m \ll n$), the goal is to design a query strategy $Q: D_U \rightarrow D_L$ to keep $m$ as small as possible while ensuring a pre-defined accuracy \citep{settles2009active}. The queried samples will be manual-labeled by a human expert with labels $y \in \mathcal{Y}$. The training objective is defined as:
\begin{equation} \label{eqn:al}
   m = \operatorname*{argmin}_{(x,y) \in D_U, (\text{x}, \text{y}) \in D_L} L_{sup}(f(\text{x}), \text{y}) +  L_{sup}(f(x), y),
\end{equation}
where $f: X \rightarrow Y \text{ or } \mathcal{X} \rightarrow \mathcal{Y}$ is the learned model, such as a deep learning model. $L(\cdot)$ denotes the loss function. Therefore, the query strategy $Q$ in active learning is crucial to reduce labeling costs. Following the literature \citep{settles2009active}, we classify active learning into three categories based on the common query strategies: \textit{uncertainty-based active learning}, \textit{Bayesian learning-based active learning}, and \textit{automated active learning}.

\begin{figure}[!ht]
  \centering
\includegraphics[width=0.49\textwidth]{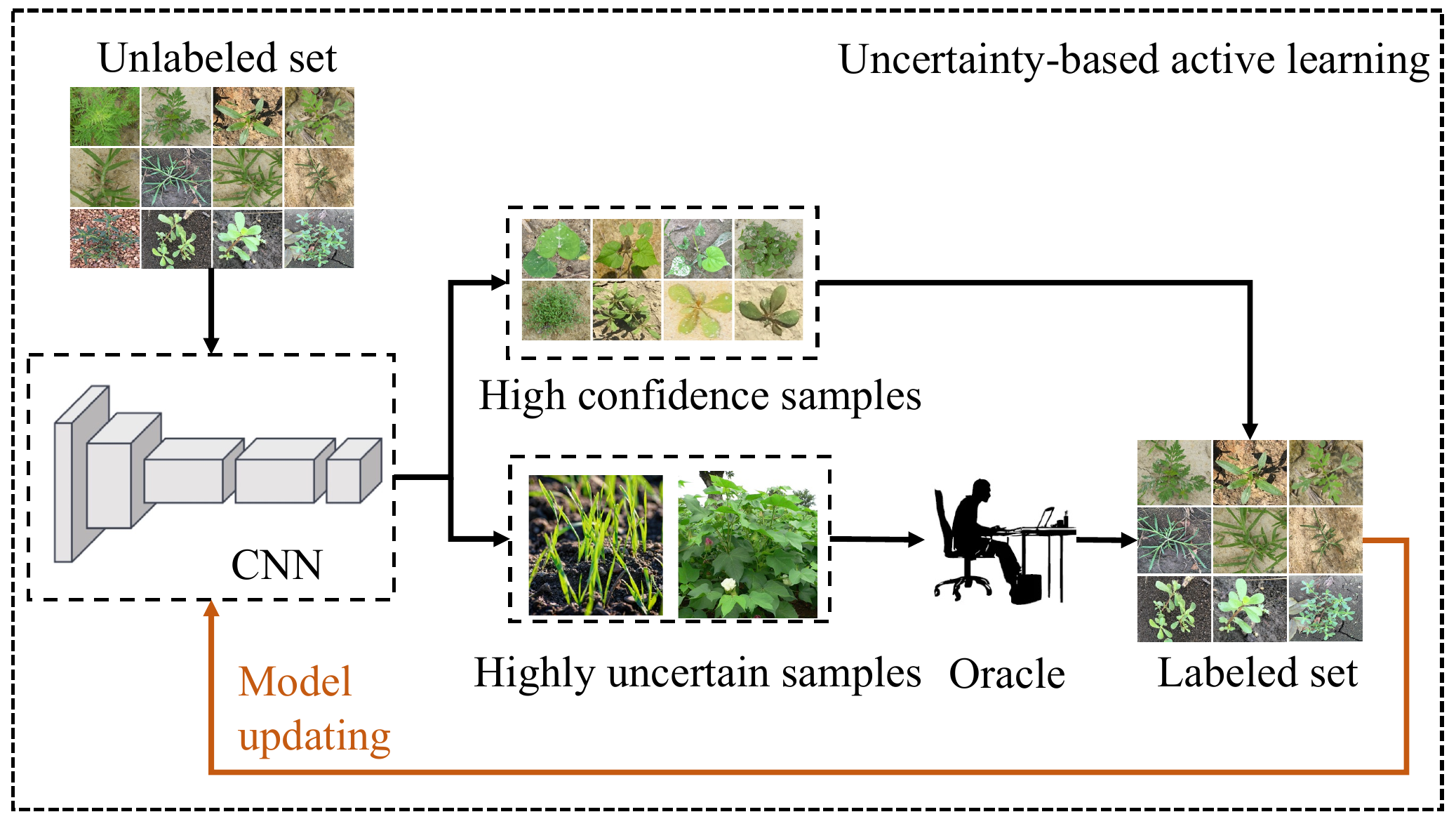}
  \caption{The framework of the uncertainty-based query strategy.}
  \label{fig:uncertian_al}
  \vspace{-10pt}
\end{figure}

\textbf{\emph{Uncertainty-based active learning.}} Uncertainty sampling is the most common and widely used query strategy. In this framework (see Fig.~\ref{fig:uncertian_al}), the instances with the most uncertainty are selected and queried. Confidence level, margin, and entropy are the three most common methods to measure the uncertainty of a sample. For instance, the least certain samples with the smallest predicted probability are chosen and labeled by an expert in \citep{lewis1994heterogeneous}. In \cite{scheffer2001active}, the active learner selects the top-K samples with the smallest margin $\mathcal{M}$ (\textit{i.e.,} most uncertain), where the margin $\mathcal{M} = P(y_1 | x) - P(y_2 | x)$ is defined as the difference between the highest predicted probability and the second highest predicted probability of a sample using the trained model. In \cite{settles2009active}, information entropy is applied as the uncertainty measure. For a $k$-class classification task, the information entropy $\mathcal{E}$ is defined as:
\begin{equation} 
\label{eqn:entropy}
   \mathcal{E}(x) = -\sum_{i=1}^k P(y_i|x) \log (P(y_i|x)),
\end{equation}
where $P(y_i|x)$ denotes the predicted probability for sample $x$. The top-K samples with the largest entropy are selected and queried. For more uncertainty-based active learning methods, readers are referred to \citep{settles2009active, 7bb36d742bf34b9d854aa0778ec47903}.

\textbf{\emph{Bayesian learning-based active learning. }} 
In \cite{houlsby2011bayesian}, Bayesian active learning by disagreement (BALD) was proposed to select samples that maximize the mutual information between model parameters and model predictions. The higher the mutual information value, the higher the uncertainty of the sample. In \cite{gal2017deep}, the authors extended BALD to the deep Bayesian active learning (DBAL) that combines Bayesian convolutional neural networks \citep{gal2015bayesian} and active learning framework to process high-dimensional image data. Evaluated on MNIST \citep{deng2012mnist} dataset, DBAL achieved 5\% test error with only 295 labeled samples and 1.64\% test error with an extra 705 labeled images, outperforming random sampling approach with 5\% test error using 835 labeled samples.

\textbf{\emph{Automated active learning. }} 
The design of previously mentioned active learning algorithms often requires substantial research experience, which hinders the adoption from users without adequate technical understanding. Therefore, it is beneficial to contemplate the automation of the design of active learning query strategies. In \cite{haussmann2019deep}, the acquisition function was replaced by a policy Bayesian neural network (BNN). The policy BNN selects the optimal samples and gets feedback from the oracle to adjust its acquisition mechanism in a reinforcement learning way \citep{kaelbling1996reinforcement}, which is often referred to the reinforcement active learning (RAL). Instead of focusing on querying mechanisms, neural architecture search \citep{ren2021comprehensive} (NAS) was employed in \citep{geifman2019deep} to automatically search for the most effective network architecture from a limited set of candidate architectures in every active learning round. The designed algorithm can then be integrated with the aforementioned querying strategies. 


\subsubsection{Semi-supervised learning} \label{sec:semi}
Semi-supervised learning aims to utilize unlabeled samples to facilitate learning without human intervention as used in active learning. Following \cite{van2020survey}, semi-supervised learning can be mainly categorized into two categories: \textit{inductive methods} and \textit{transductive methods}, based on how the unlabeled samples are incorporated. Inductive methods extend supervised algorithms with unlabeled data to jointly train the prediction model, whereas transductive techniques are typically graph-based approaches that directly produce the predictions for the unlabeled data.

\begin{figure}[!ht]
  \centering
\includegraphics[width=0.49\textwidth]{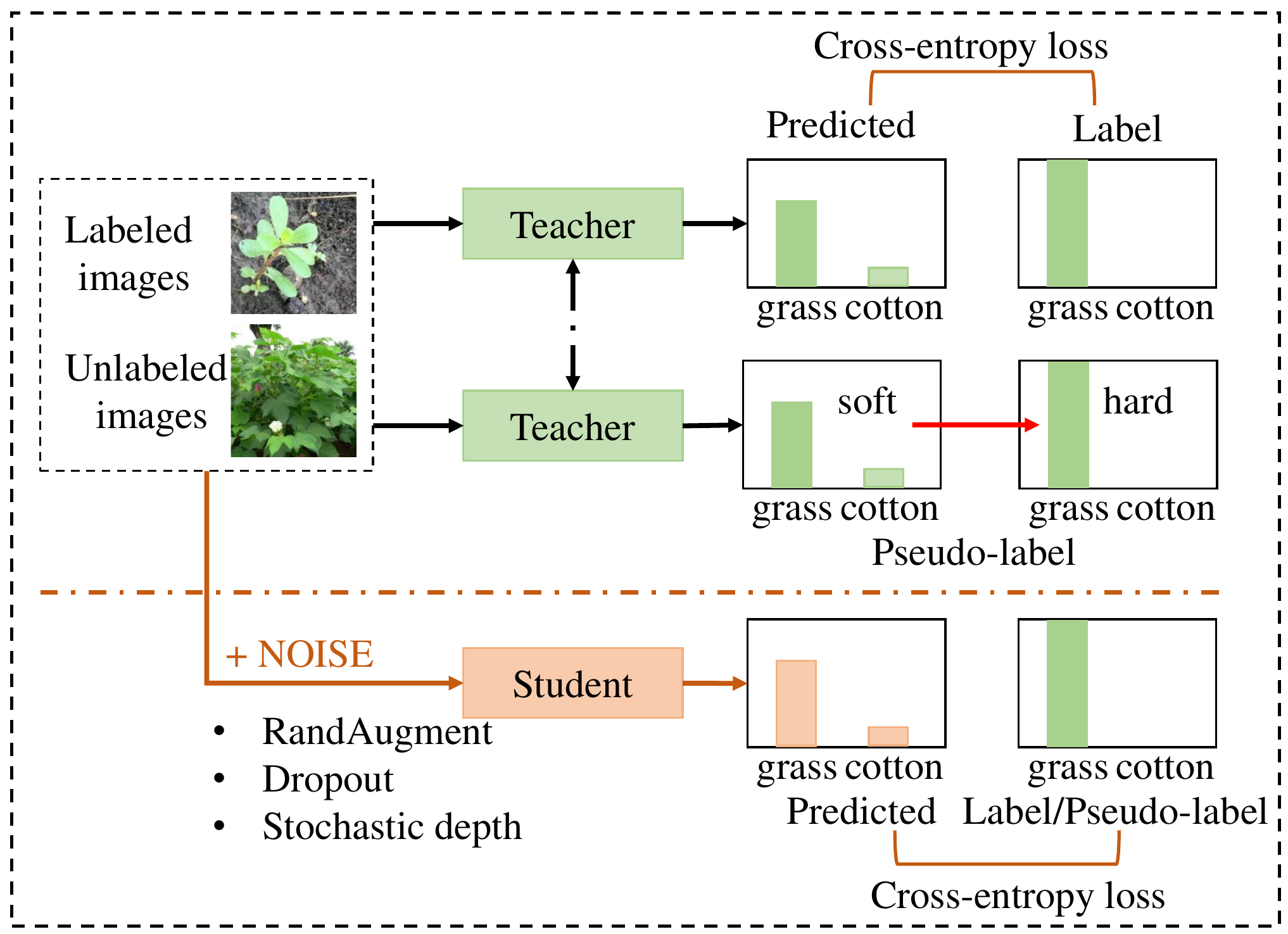}
  \caption{The framework of the ``noisy student'' algorithm \citep{xie2020self}.}
  \label{fig:noisystudent}
  \vspace{-10pt}
\end{figure}

\textbf{\emph{Inductive semi-supervised methods. }} \textit{Self-training} 
(\textit{a.k.a. self-learning}) \citep{yarowsky1995unsupervised} methods are the most basic semi-supervised learning approaches. First, a supervised model is trained only on the labeled samples, then the obtained model is applied to generate predictions for the unlabeled samples (also known as pseudo-labeled data). The most confident pseudo-labeled samples and the original labeled samples are then jointly used to re-train the supervised model, and the process often repeats until a satisfactory performance is achieved. Therefore, self-training methods are versatile and can be integrated with any supervised learning-based approaches.
In \cite{lee2013pseudo}, the pseudo-label approach was proposed for image classification and evaluated on the MNIST dataset \citep{deng2012mnist} with promising performance demonstrated. To improve the reliability of pseudo-labels, the weighting of pseudo-labeled samples was generally increased over time during the training process. In \cite{xie2020self}, the ``noisy student'' algorithm was proposed for image classification tasks. As shown in Fig.~\ref{fig:noisystudent}, in the framework a teacher model was first trained on the labeled samples in a supervised way. The trained teacher model then generated pseudo labels on the unlabeled samples. A student model (large or equal size as the teacher model) was trained with both the labeled and pseudo-labeled samples. During the training of the student model, input and model noise, such as stochastic depth, dropout, and data augmentation, were applied to help the student model generalize better than the teacher model. However, the teacher model was trained without noise to provide accurate pseudo labels. 
In \cite{liu2021unbiased}, the authors applied the teacher-student framework for semi-supervised object detection. The inputs were augmented with weak and strong data augmentation for the teacher and student models, respectively. To combat class imbalance issues, focal loss \citep{lin2017focal} was applied and the teacher was progressively updated with the student via the exponential moving average (EMA) update \citep{tarvainen2017mean}. 

\textit{Co-training} \citep{blum1998combining} (also named \textit{disagreement-based}) methods extend the self-training to two or more supervised learners and exploit disagreements among the learners to improve the performance of the machine learning model. The learners provide independent pseudo-labels for the unlabeled samples and exchange information through the unlabeled samples to improve their performance. For example, in \citep{zhou2011semi, zhou2012ensemble}, multiple learners were incorporated into an ensemble for better generalization. The ensemble learning \citep{dong2020survey} approach combined the predictions of multiple base learners and makes a final decision based on their combined output. 

\textit{Intrinsically semi-supervised} methods \citep{van2020survey} are another type of semi-supervised learning approach that directly incorporate unlabeled samples into the objective function without any intermediate steps or supervised base learner. Among the methods, the most widely used is semi-supervised support vector machines \citep{vapnik1998statistical} (S3VMs). It tries to identify a classification boundary in a low-density area \citep{ben2009learning} that correctly classifies the labeled samples with as few unlabeled samples violating the classification margin as possible. Subsequently, \cite{li2013convex, chapelle2008optimization} were further proposed for improving the optimization efficiency. For instance, \cite{li2013convex} proposed the WELLSVM algorithm as a solution to address the issue of poor scalability in semi-supervised learning, which can lead to the occurrence of local minimum problems.


\textbf{\emph{Transductive semi-supervised methods.}} 
The aforementioned inductive semi-supervised approaches use both labeled and unlabeled samples to construct a model and provide predictions for the entire data, while transductive semi-supervised methods are only generating predictions for the unlabeled samples. Transductive methods typically construct a graph $\mathcal{G} = (\text{\larger[2]$\nu$}, \text{\larger[2]$\varepsilon$})$ over all data samples (\textit{i.e.,} labeled and unlabeled), where each node $\nu_i \in \text{\larger[2]$\nu$}$ represents a training sample and the edge  $\text{$\varepsilon_{ij}$} \in \text{\larger[2]$\varepsilon$}, i \neq j$ corresponds to the relation (\textit{e.g.,} distance or similarity) between the sample $i$ and $j$. Through the graph,  data points with small dissimilarities are viewed as ``connected'', thus the label information of labeled samples can be propagated to the unlabeled samples through the edge connections. More specifically, the labeled data samples are used as the initial labels for the labeled data points in the graph, and these labels are propagated to the unlabeled data points by iteratively updating the label of each data point based on the labels of its neighbors in the graph. This process continues until the labels converge or a stopping criterion is met. The resulting labels for the unlabeled data points can be used to make predictions or classifications for new data points. Graph construction and inference over graphs are two key aspects of graph-based transductive methods. 

Graph construction is to form a graph structure that captures the similarities among data points, which is characterized by an adjacency matrix and an edge attribute matrix. The adjacency matrix builds the connections between nodes, while the edge attribute matrix determines the weights (i.e., distance or similarity) for the edges in the graph. $\epsilon$-neighborhood \citep{blum2001learning}, $k$-nearest neighbors \citep{blum2001learning}, and $b$-matching \citep{jebara2009graph} are the three most common approaches to build the adjacency matrix. Specifically, $\epsilon$-neighborhood \citep{blum2001learning} connects data samples with a distance (e.g., Euclidean distance) below a pre-defined threshold $\epsilon$. Obviously, the performance is heavily dependent on the choice of $\epsilon$, which largely limits its applications in real-world applications. On the other hand, $k$-nearest neighbors methods \citep{blum2001learning, maier2008influence}, as the most common graph construction method, connect each node to its $k$ nearest neighbors based on some distance measure (e.g., Euclidean distance). Both $\epsilon$-neighborhood and $k$-nearest neighbors methods determine the node's neighbors for each node independently from the perspective of local observations, which often leads to sub-optimal solutions \citep{jebara2009graph}. 
To address this issue, the $b$-matching method proposed by \cite{jebara2009graph} constructs the graph via optimizing a global objective, ensuring each node has the same number of neighbors and edge connections to enforce the regularity of the graph. As for graph weighting, it refers to the process of assigning weights to the graph edges. Gaussian edge weighting \citep{de2013influence} is one of the most common weighting approaches that use Gaussian kernel as the similarity measure of edge connections. 

In practice, graph-based transductive semi-supervised methods suffer severe scalability issues due to costly computational complexity during graph construction and inference \citep{liu2012robust, chong2020graph}. Also, they are difficult to accommodate new samples without graph reconstruction. Recently, \cite{liu2014random} and  \cite{zhang2017mixup} tackled the scalability problem by constructing smaller subgraphs so that graph inference can be executed efficiently. For more graph-based semi-supervised learning methods, \citep{chong2020graph} is referred for further reading.


\begin{figure}[!ht]
  \centering
\includegraphics[width=0.48\textwidth]{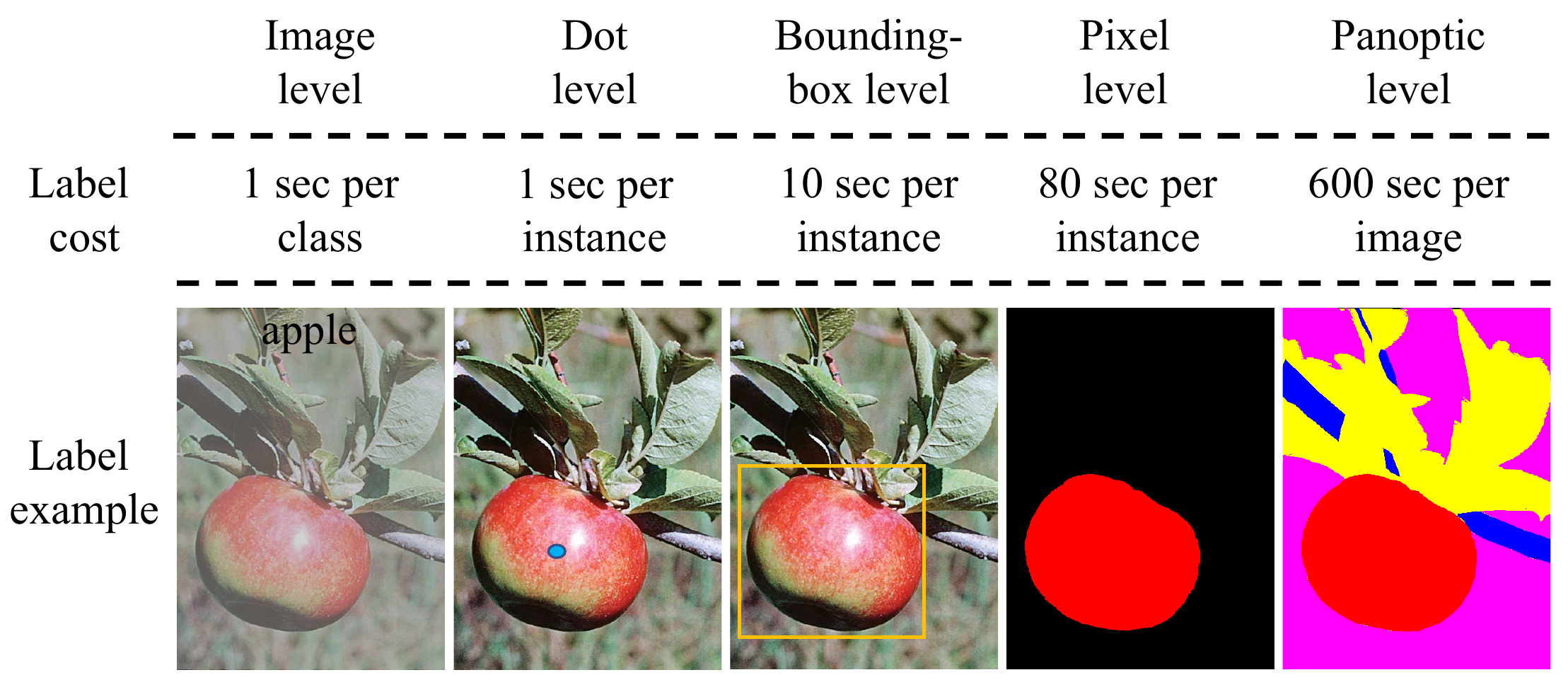}
  \caption{Examples of label costs of different annotation types.}
  \label{fig:annotation_time}
  \vspace{-10pt}
\end{figure}

\subsubsection{Weakly supervised learning}
\label{sec:wsl}
Weakly supervised learning is often applied in scenarios where only partial information (e.g., coarse-grained, incomplete, or noisy labels) is provided. This is particularly useful in scenarios where it is expensive or impractical to obtain densely labeled data, such as in medical imaging or satellite imagery analysis. For example, in the problem of fruit segmentation and tracking \citep{ciarfuglia2023weakly}, it is extremely time-consuming and cost-intensive to obtain fine-grained annotations (i.e., dense pixel-level annotations). Instead, image-level or bounding-box annotations are cheaper to obtain (Fig.~\ref{fig:annotation_time}).

To bridge the gap between weak supervision and dense supervision signals, some heuristic priors are leveraged: 1) pixels of the same classes often share common features like color, brightness, and texture; 2) semantic relationships remain among pixels belonging to objects of the same category across distinct images \citep{shen2022survey}. \textit{Multi-instance learning} (MIL) and \textit{class activation mapping} (CAM) methods are two representative weakly supervised learning approaches, which are detailed as follows. 

\begin{figure}[!ht]
  \centering
\includegraphics[width=0.4\textwidth]{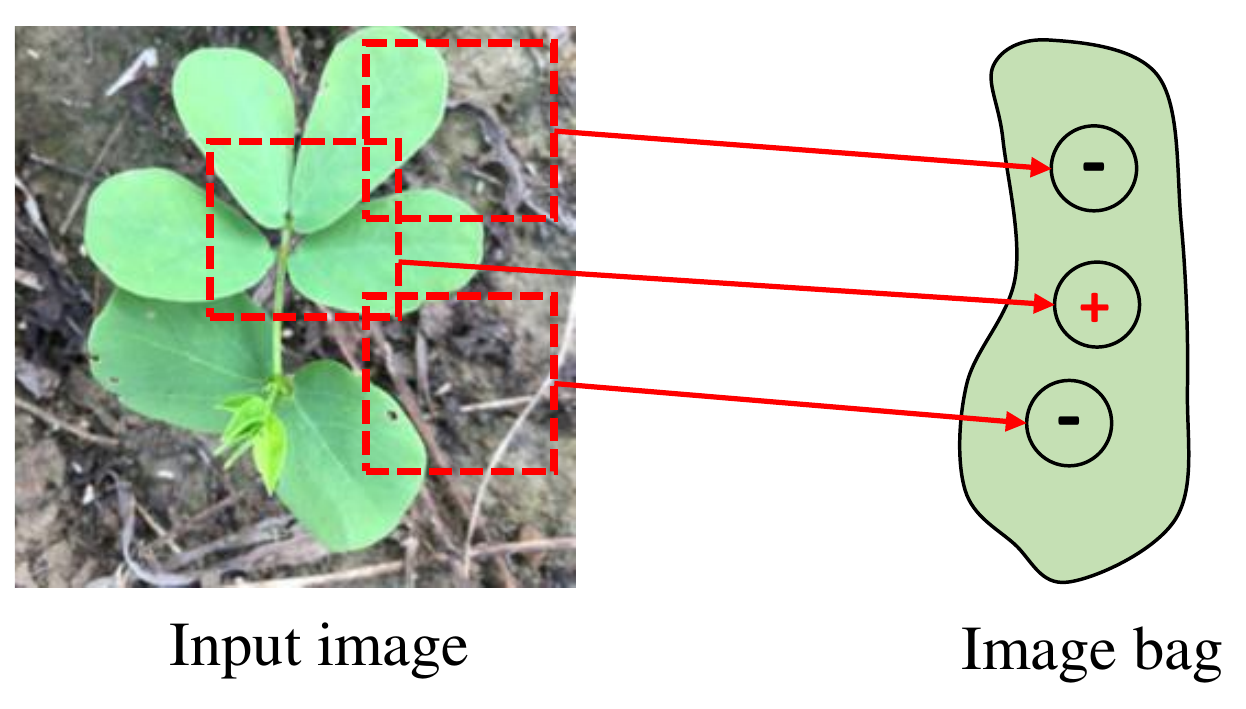}
  \caption{The instance and bag generation process of multi-instance learning (MIL). The bags are instances that are grouped in sets.  ``+'' and ``-'' represent positive and negative instances, respectively.}
  \label{fig:mil}
  \vspace{-10pt}
\end{figure}

\textbf{\emph{Multi-instance learning (MIL)}} \citep{foulds2010review, carbonneau2018multiple} has attracted increased research attention recently to alleviate growing labeling demands. In MIL, training instances are grouped in sets, called bags, and ground-truth labels are only available for the entire sets instead of individual bags, which naturally fits various applications where only weak supervision labels (e.g., image-level labels in Fig.~\ref{fig:annotation_time}) are given. The process of instance and bag generation is shown in Fig.~\ref{fig:mil}. The predictions can be generated at the bag level or instance level. For example, mi-SVM and MI-SVM MIL algorithms were proposed by \cite{andrews2002support} for instance-level and bag-level classification, respectively. Conventional machine learning-based MIL approaches struggle with high-dimensional visual inputs \citep{wu2015deep}. However, researchers have recently turned to the study of weakly supervised learning, utilizing deep representations to learn features \citep{wu2015deep, ilse2018attention}.  Refer to \citep{carbonneau2018multiple} for more MIL methods.

\begin{figure}[!ht]
  \centering
\includegraphics[width=0.45\textwidth]{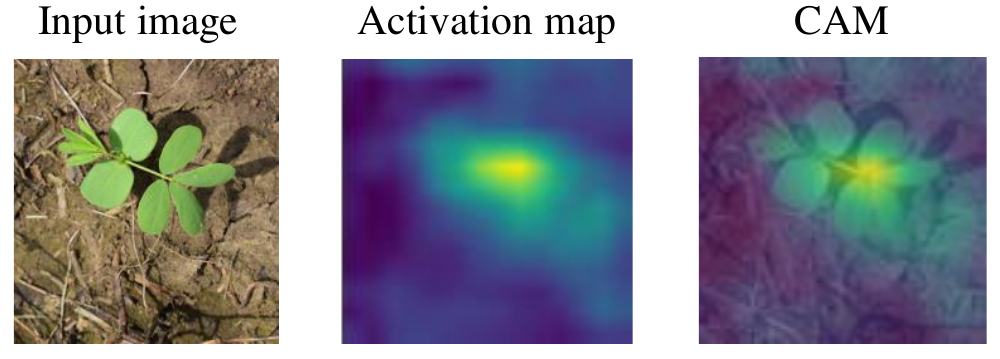}
  \caption{Visualization of applying Class activation mapping (CAM) \citep{zhou2016learning} on a ``Sicklepod'' weed image.}
  \label{fig:CAM}
  \vspace{-10pt}
\end{figure}

\begin{figure}[!ht]
  \centering
\includegraphics[width=0.46\textwidth]{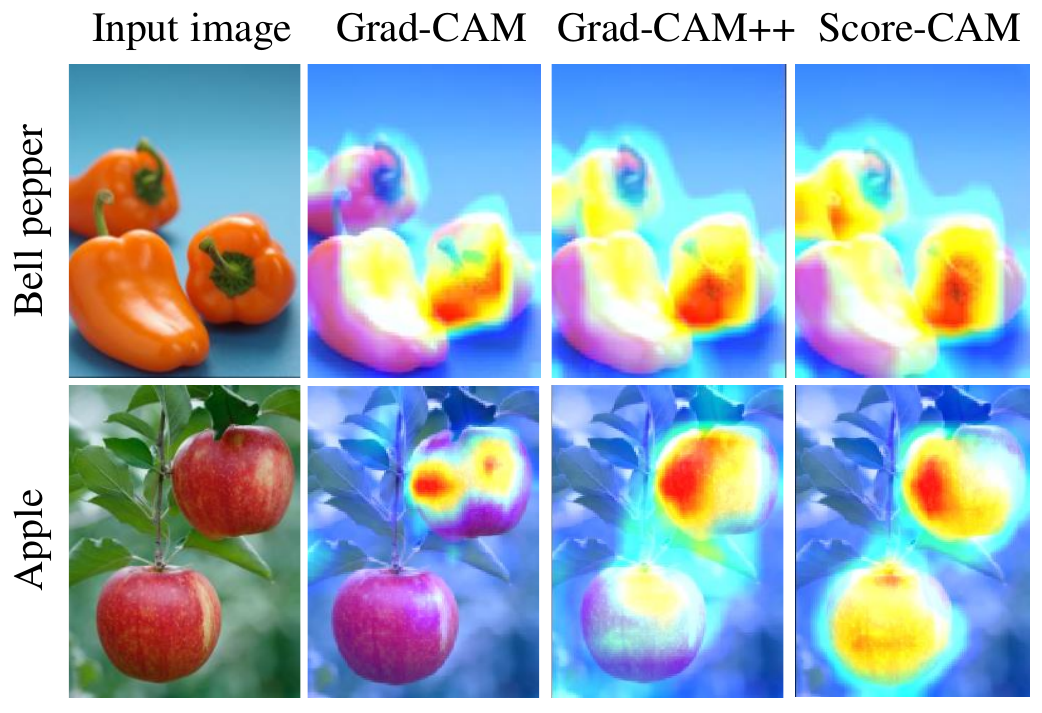}
  \caption{Visualization of Grad-CAM \citep{selvaraju2017grad}, Grad-CAM++ \citep{chattopadhay2018grad}, and Score-CAM \citep{wang2020score} on two input images. Score-CAM shows a higher concentration at the relevant object.}
  \label{fig:CAM_compare}
  \vspace{-10pt}
\end{figure}

The \textbf{\emph{Class activation mapping (CAM)}} technique was proposed in \cite{zhou2016learning} for discriminative object classification and localization with only image-level labels. As shown in Fig.~\ref{fig:CAM}, CAM highlights the class-discriminative regions reflecting important image regions, which is generated by performing global average pooling on the convolutional layers for the output layers and then mapping the predicted class scores back to  the previous convolutional layers by taking a weighted linear summation. However, standard CAM is generally architecture-sensitive and only works for particular kinds of convolutional neural network (CNN) architectures/models with fully-connected layers due to the direct connection between the feature maps generated by the global average pooling and output layers (i.e., softmax layers). To this end, 
gradient-weighted CAM \citep{selvaraju2017grad} (Grad-CAM) was developed to address this issue by sending the gradient signals of class information back to the last convolutional layer without any modifications on the networks. To alleviate the performance degradation when inferring multiple objects within the same class, Grad-CAM++ \citep{chattopadhay2018grad} extended Grad-CAM with a more general formulation by introducing a pixel-wise weighting scheme to capture the spatial importance of the regions on the convolutional feature maps. While gradient-based CAM approaches have achieved promising progress, Score-CAM \citep{wang2020score} argued that these approaches often suffer from gradient vanishing and false confidence issues, resulting in unsatisfactory performance on specific tasks. \cite{wang2020score} replaced the gradient methods \citep{chattopadhay2018grad} by taking a linear combination of score-based weights and activation maps. Fig.~\ref{fig:CAM_compare} shows the visualization comparison of Grad-CAM \citep{selvaraju2017grad}, Grad-CAM++ \citep{chattopadhay2018grad}, and Score-CAM \citep{wang2020score} on two input images. It is obvious that Score-CAM shows a higher concentration at the relevant objects.

\subsection{No Supervision}
The objective of unsupervised learning is to train a representation from unlabeled data that can be applied to future tasks. This can be achieved through various methods such as self-supervised learning, unsupervised representation learning, and generative models like auto-encoders and GANs \citep{qi2020small}. These techniques allow for efficient generalization and improved performance in downstream tasks. 
Specifically, self-supervised learning focuses on predicting implicit features in the data like spatial relationships or transformations, while unsupervised representation learning creates a condensed and informative representation of the data without explicit guidance. Generative models like auto-encoders learn to compress the input data into a latent space representation, and then reconstruct the original data from the compressed representation. GANs, on the other hand, learn to generate new data by training a generator network to produce data that is similar to the input data, and a discriminator network to distinguish between real and fake data.
Although generative models have shown promising results in unsupervised learning, there are other review papers \citep{lu2022generative, qi2020small} that cover this topic in detail. We will thus not discuss it in detail in this survey.

\subsubsection{Self-supervised learning} \label{sec:selfsl}
\begin{figure}[!ht]
  \centering
\includegraphics[width=0.45\textwidth]{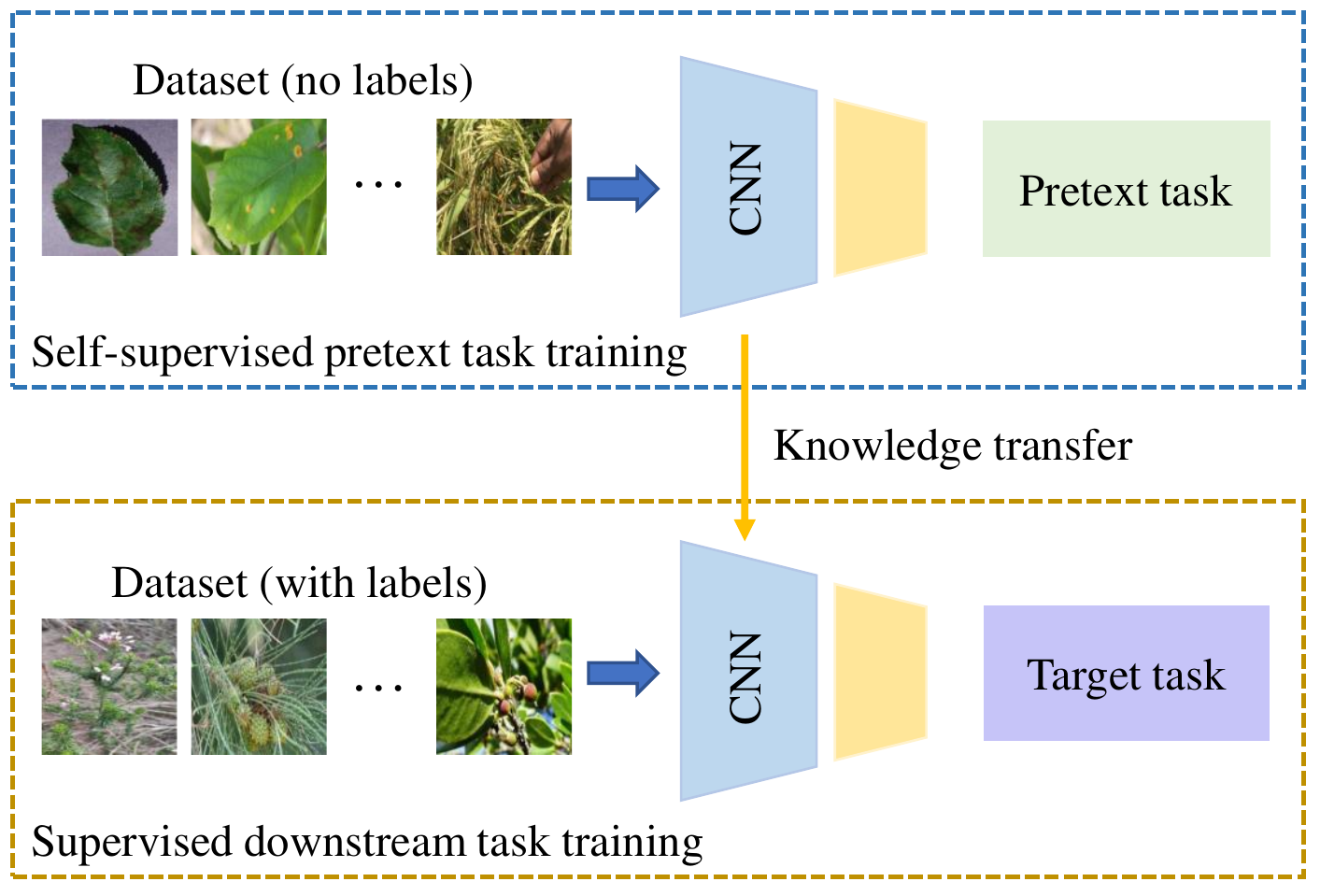}
  \caption{The pipeline of self-supervised learning.}
  \label{fig:self-supervised}
  \vspace{-10pt}
\end{figure}

Self-supervised learning \citep{jing2020self, schmarje2021survey} is a branch of  unsupervised learning approaches that aims to train ML/DL models with large-scale unlabeled data without any human annotations. Fig.~\ref{fig:self-supervised} shows the general framework of self-supervised learning approaches. In the first stage (i.e., self-supervised pretext task training), the ML/DL model (e.g., convolutional neural networks (CNN)) is explicitly trained on the unlabeled dataset to learn data representations with automatically generated pseudo labels based on data attributes such as spatial proximity, colorization, and pixel intensities \citep{jing2020self}. Since the pseudo labels are generated automatically without any human annotation efforts (see next paragraph for more details), very large-scale datasets are typically used for the self-supervised learning stage. For example, for general computer vision tasks, ImageNet \citep{deng2009imagenet} and Microsoft COCO \citep{lin2014microsoft} are often used as the pretext tasks, and large-scale image datasets, e.g., PlantCLEF2022 \citep{goeau2022overview}, are often served as the pretext datasets for the agricultural applications \citep{xu2022transfer}. 
After the self-supervised training is finished, the trained model is fine-tuned on a small number of labeled samples (targeted dataset) through knowledge transfer to overcome overfitting \citep{ying2019overview} and improve the model performance. 

Following the literature, self-supervised learning is typically categorized into \textit{generative}, \textit{contrastive}, and \textit{generative-contrastive} (\textit{adversarial}) \citep{jing2020self, liu2021self}. In this survey, we will only focus on contrastive learning methods, which are the more commonly used in agricultural applications.

\begin{figure}[!ht]
  \centering
\includegraphics[width=0.49\textwidth]{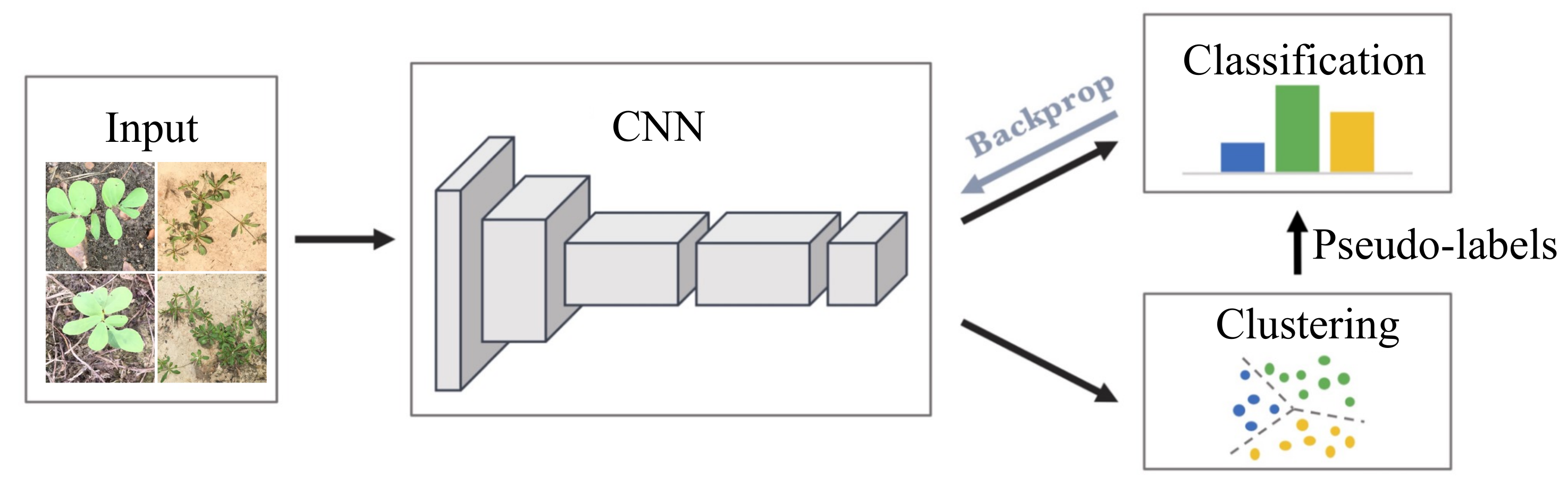}
  \caption{The framework of DeepCluster algorithm. Deep features are clustered iteratively and the cluster assignments are used as pseudo-labels to learn the parameters of CNN \citep{caron2018deep}.}
  \label{fig:deepclustering}
  \vspace{-10pt}
\end{figure}

In supervised learning, samples are grouped via label supervision. In self-supervised learning, however, no labels are available for supervised learning. To address this issue, DeepCluster \citep{caron2018deep} (see Fig.~\ref{fig:deepclustering}), a clustering algorithm was first employed to produce pseudo labels by grouping similar images in the embedding space generated by a convolutional neural network (CNN). A classifier (i.e., discriminator) was then trained to tell whether two input samples are from the same cluster, and the gradients were back-propagated to train the CNN. To learn the semantic meaning of the images, the CNN was trained to capture the similarities within the same image class while also detecting the differences between different classes via the cross-entropy discriminative loss. There are also other works employing clustering methods during the pretext task training, such as \cite{yang2016joint, xie2016unsupervised, noroozi2018boosting, zhuang2019local}. Typically in such methods, input images are first encoded into the embedding space, followed by clustering these embedded features into distinct groups based on a distance measurement. Finally, a CNN is trained to differentiate between images from the same cluster and those from different clusters. 

\begin{figure}[!ht]
  \centering
\includegraphics[width=0.48\textwidth]{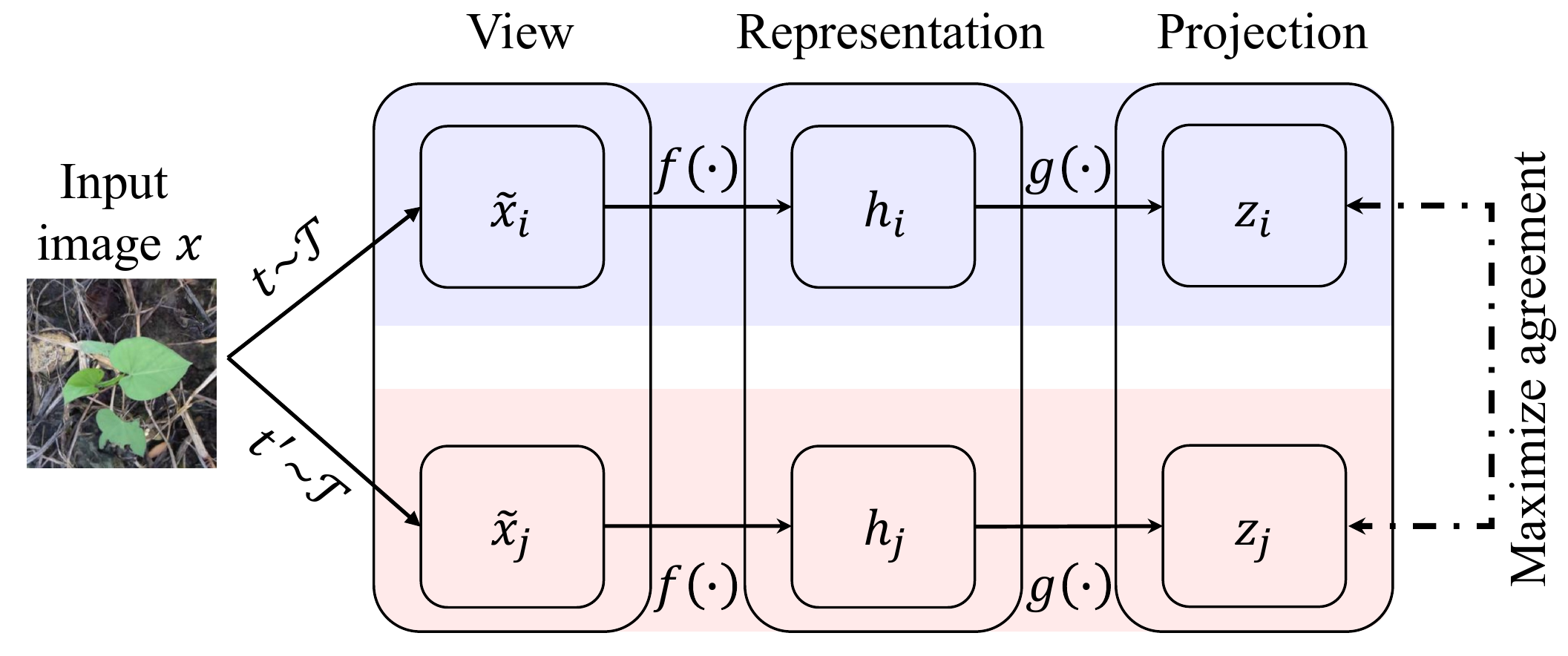}
  \caption{The framework of SimCLR  \cite{chen2020simple} algorithm.}
  \label{fig:SimCLR}
  \vspace{-10pt}
\end{figure}

\begin{figure}[!ht]
  \centering
\includegraphics[width=0.49\textwidth]{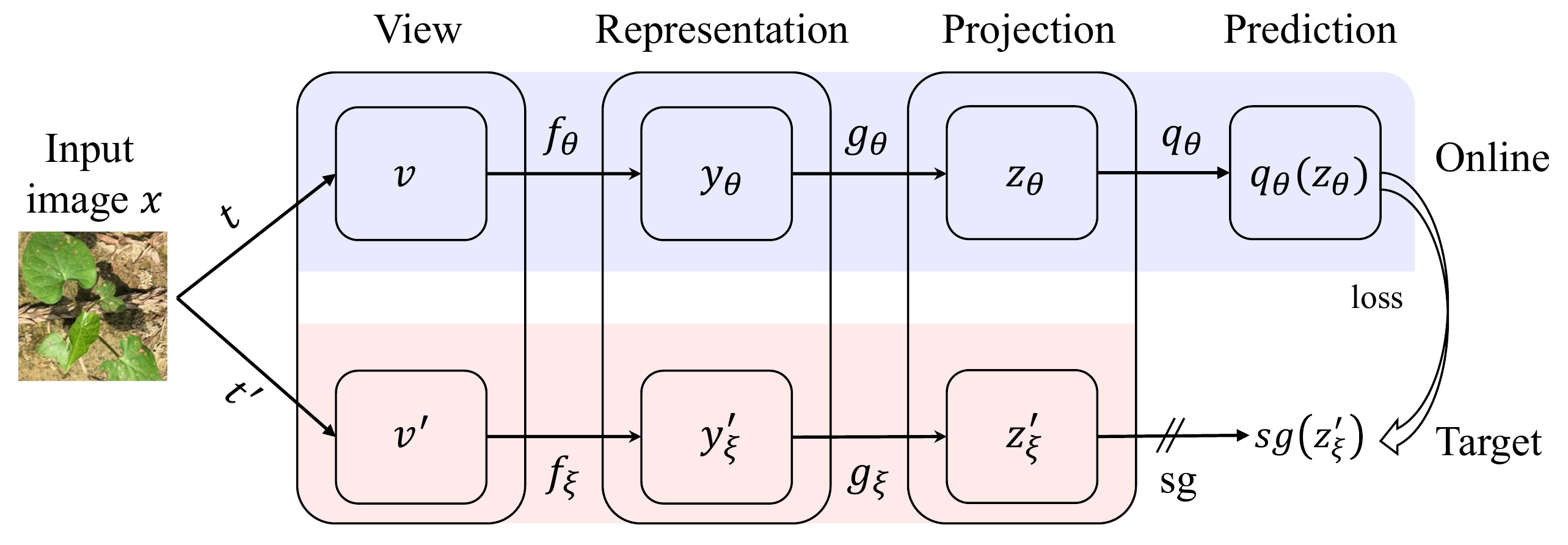}
  \caption{The framework of BYOL algorithm \citep{grill2020bootstrap}.}
  \label{fig:BYOL}
  \vspace{-10pt}
\end{figure}

Contrastive learning via different views over image pairs is also explored in \cite{tian2020contrastive, caron2020unsupervised}, in which different views of the same image are treated as positive samples, while different images are considered as negative ones. Siamese network structures \citep{chicco2021siamese} are widely adopted in these approaches by maximizing the similarity between two augmentations of one image. For instance,
SimCLR \cite{chen2020simple} utilized a contrastive loss in the latent space to learn the representation of visual inputs. As shown in Fig.~\ref{fig:SimCLR}, two correlated views of each data sample were obtained by randomly applying two data augmentation operators ($t\sim\mathcal{T}$ and $t' \sim \mathcal{T}$) on the input sample. Then, a base encoder network $f(\cdot)$ and a projection head $g(\cdot)$ were trained using a contrastive loss to maximize the agreement between different augmented views of the same sample to enhance the quality of the learned features. Once training was finished, the projection head $g(\cdot)$ was discarded, and the encoder $f(\cdot)$ and the learned representation $h$ were utilized for downstream tasks.
In another influential self-supervised learning framework \cite{grill2020bootstrap}, BYOL (\underline{B}ootstrap \underline{Y}our \underline{O}wn \underline{L}atent), two neural networks, known as the online and target networks, collaborated and acquired knowledge from each other without using negative samples. As shown in Fig.~\ref{fig:BYOL}, BYOL cast the prediction problem directly in the representation space, where the online network learned to predict the target network's representation of the same image from different views. The target network was constructed with the same architecture as the online network, but its parameters were updated with the exponential moving average (EMA) strategy. Once the training was completed, only $f_{\theta}$ and $y_{\theta}$ were kept for the image representation. In SimSiam \citep{chen2021exploring}, the authors showed that simple Siamese networks can also learn meaningful representations without negative image pairs, large image batches, or momentum encoders. With a stop-gradient operation, SimSiam achieved a faster convergence speed than SimCLR, SwAV, and BYOL even with smaller batch sizes.  

\begin{figure}[!ht]
  \centering
\includegraphics[width=0.48\textwidth]{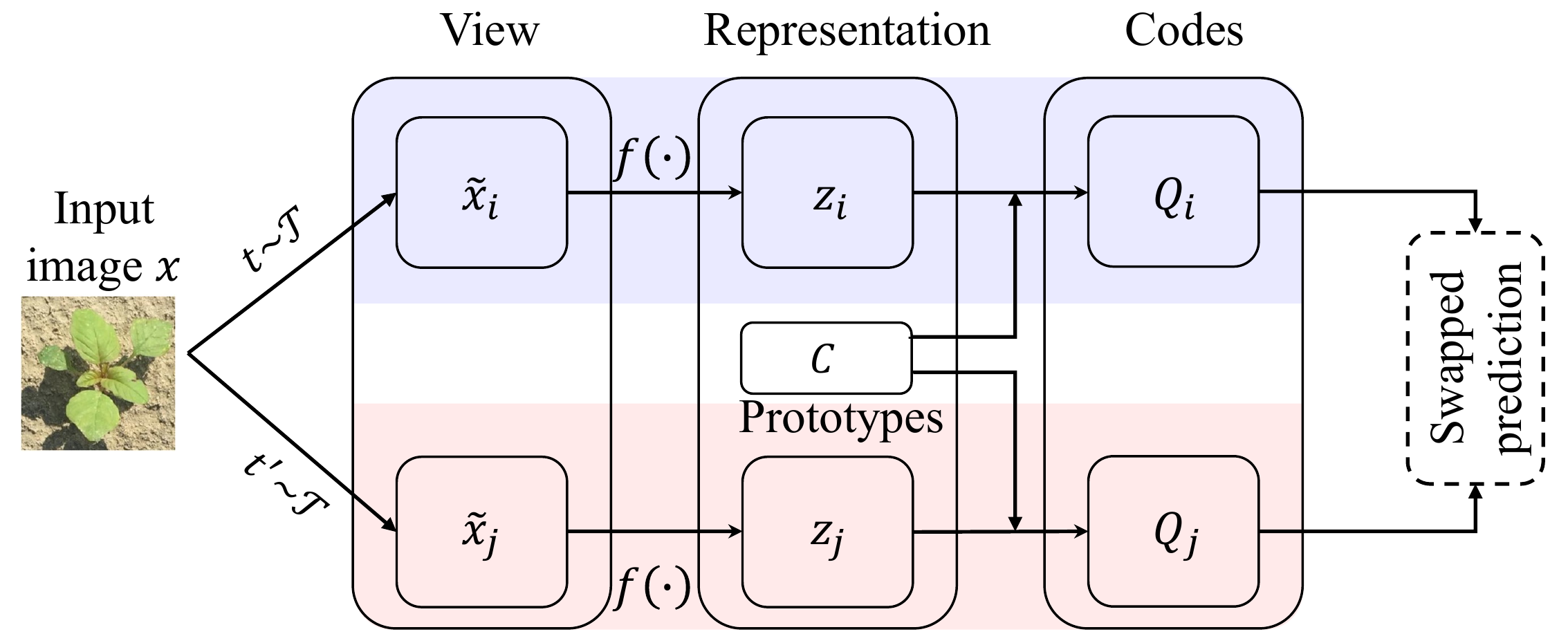}
  \caption{The framework of SwAV algorithm \citep{caron2020unsupervised}.}
  \label{fig:SwAV}
  \vspace{-10pt}
\end{figure}

Despite decent progress on cluster discrimination-based contrastive learning \citep{caron2018deep, zhuang2019local}, the  clustering stage is generally slow with  poor performing as compared to later multi-view contrastive learning approaches. In light of these problems, SwAV (\underline{sw}apping \underline{a}ssignments between multiple \underline{v}iews, \cite{caron2020unsupervised}) addressed these issues by combining online clustering ideas and multi-view data augmentation techniques into a cluster discrimination approaches. Instead of comparing features directly as in contrastive learning, SwAV (Fig.~\ref{fig:SwAV}) utilized a ``swapped'' prediction mechanism, in which the code of a view from the representation of another view was predicted. Experimental results showed that SwAV achieved state-of-the-art performance and surpassed the supervised learning approach on all the downstream tasks through knowledge transfer.

\subsubsection{Unsupervised representation learning}
Clustering is another common technique that involves categorizing similar sets of data into distinct clusters (groups) based on some similarity measures (e.g., Euclidean distance). This powerful method utilizes the attributes of the data to group it, making it a widely used approach in a range of fields such as machine learning, image processing, and video processing \citep{min2018survey}. For example, $k$-means \citep{hartigan1979algorithm} is a widely used clustering algorithm that partitions a dataset into $k$ clusters based on the similarity of the data points by iteratively assigning each data point to the closest cluster centroid, and then updating the centroid by taking the mean of all the data points assigned to it. The algorithm terminates when the cluster assignments no longer change, or a maximum number of iterations is reached. Although $k$-means is simple to implement and scalable to large datasets, it is sensitive to the initial choice of centroids and may not perform well on datasets with complex structures \citep{arthur2007k}. To address the above issues, $k$-means++ \citep{arthur2007k} improved $k$-means by generating the initial cluster centroids by a more sophisticated seeding procedure to spread the initial centroids out across the dataset well and reduce the likelihood of getting stuck in a local optimum.

\begin{figure}[!ht]
  \centering
\includegraphics[width=0.48\textwidth]{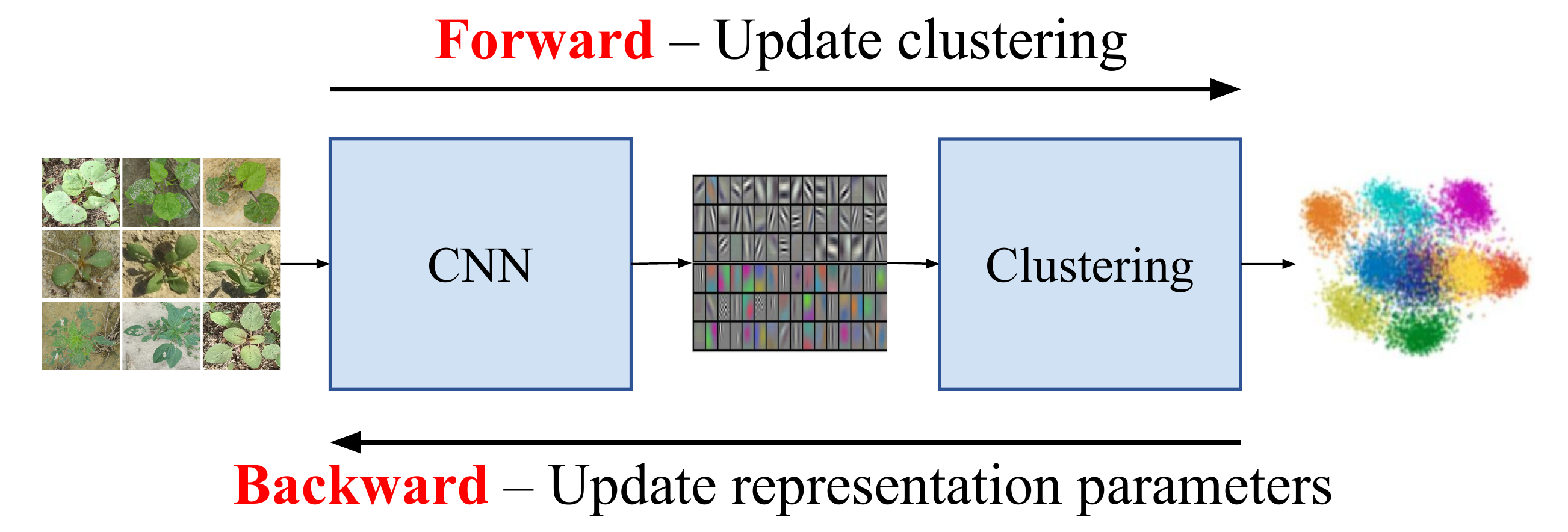}
  \caption{The framework of JULE algorithm \citep{yang2016joint}.}
  \label{fig:JULE}
  \vspace{-10pt}
\end{figure}

Recently, there has been growing interest in using deep learning-based clustering approaches. Compared to traditional clustering approaches, these methods perform better in processing high-dimensional and heterogeneous data and capture non-linear relationships between data points by leveraging deep neural networks. In contrast to traditional clustering methods, which often rely on handcrafted features and assumptions about data distributions, deep clustering approaches can automatically learn useful representations of the data from raw inputs \citep{aljalbout2018clustering}. For example, JULE \citep{yang2016joint}  proposed a recurrent unsupervised learning framework for simultaneously learning deep feature representations and clustering image data. As shown in Fig.~\ref{fig:JULE}, in the forward pass, image clustering was conducted via using Agglomerative clustering \citep{gowda1978agglomerative} algorithm, while parameters of representation learning were updated through the backward process. Furthermore, a single loss function was derived to guide Agglomerative clustering and deep representation learning simultaneously, benefiting from good representations, and providing supervisory signals for clustering results.

\section{Applications of Label-Efficient Learning in Agriculture}
\label{sec:apps}
In this section, we present a review of  application studies of label-efficient learning in the agricultural field. These applications are organized into three main areas: precision agriculture, plant phenotyping, and postharvest quality assessment of agricultural products. The reviewed papers are also categorized by the degrees of required supervision defined in Section~\ref{sec:concepts}.

\subsection{Precision agriculture}
Precision agriculture, also known as precision or smart farming, aims to improve agricultural production (e.g., crops and animals) efficiency and sustainability through more precise (e.g., site-specific) and resource-efficient farming management strategies \citep{monteiro2021precision}. It leverages  advanced technologies in robots, artificial intelligence, sensors, information theory, and communication  to support critical agricultural tasks, such as plant health monitoring, crop and weed management, fruit sorting and harvesting, and animal monitoring and tracking. In the past decade, label-efficient approaches have received significant attention from the agricultural community to reduce expensive label costs and improve learning efficiency for various applications as summarized below.

 \subsubsection{Plant health}
Plant diseases or disorders caused by biotic and abiotic stressors \citep{dhaka2021survey}, such as microorganisms (e.g., viruses, bacteria, fungi), insects, and environmental factors, negatively affect crop yield and production quality \citep{zhang2021identification, lu2022generative}. Imaging technologies through the analysis of plant leaf images (e.g., RGB, NIR, and hyperspectral images taken by various cameras or unmanned aerial vehicles (UAVs)) currently serve as promising means for characterization and diagnosis of plant health conditions \citep{xu2022style, mahlein2018hyperspectral}. Recently, machine vision-based methods (e.g., CNN) have been frequently adopted by the agricultural community with promising performance demonstrated. However, expert annotations remain costly and critical challenges to develop supervised learning-based machine vision systems with a large-scale image dataset \citep{li2021semi}. To reduce the cost in labeling, label-efficient methods (Fig.~\ref{fig:app1}) have been utilized to develop machine vision systems for identifying plant diseases and other health conditions with few or no manual-labeled annotations.  We next present the label-efficient learning applications in plant health monitoring, organized based on the taxonomy described in Section 3. \vspace{10pt} \\
\underline{\textbf{Weak supervision}} \vspace{5pt} \\
Weak supervision methods, such as active learning, semi-supervised learning, and weakly supervised learning are widely adopted for enhancing the recognition of plant diseases and other health conditions to reduce labeling costs.

\textit{\textbf{Active learning. }}
In \cite{coletta2022novelty}, active learning (Section~\ref{sec:al}) was explored to leverage unlabeled data to help identify new threats (e.g., diseases or pests) appearing in eucalyptus crops with the images acquired on Unmanned Aerial Vehicle (UAV). To detect the new threats (i.e., Ceratocystis wilt, a new disease class) with just a few labeled samples, the iterative classifier (IC, \citep{coletta2019combining}) framework was employed to identify instances with the new disease. More specifically, the entropy and density-based selection (EBS) algorithm \citep{coletta2019combining} was adopted to measure the entropy of input instances, and the unlabeled instances with high uncertain labels (i.e., high entropy) were selected and labeled by a domain expert. Then, the newly labeled instances were incorporated into the training set to refine the classification model. The authors collected an aerial image dataset containing 74,199 image instances with a resolution of $4608 \times 3456$. Experimental results showed that, with only 50 labeled samples, the proposed approach was able to reduce the identification error to 8.8\% and 12.7\% with 28.3\% and 16.5\% new diseased samples, respectively. Although promising performance was achieved, the authors did not discuss the situation when there are more than one new disease classes.

\textit{\textbf{Semi-supervised learning. }}
Semi-supervised learning (Section~\ref{sec:semi}) has also been applied to improve plant disease identification performance by employing large amounts of unlabeled data. 
For example, the pseudo-label approach (Section~\ref{sec:semi}) was employed in \cite{amorim2019semi} to utilize unlabeled samples for soybean leaf and herbivorous pest identification. Firstly, three CNNs (i.e., Inception-V3, Resnet-50, and VGG19 \citep{simonyan2014very}) were pre-trained on the ImageNet \citep{deng2009imagenet} dataset and then transferred to their own datasets through transfer learning \citep{zhuang2020comprehensive}. The unlabeled samples were pseudo-labeled by five classical semi-supervised methods, including Transductive Support Vector Machines (TSVM) \citep{joachims1999transductive}) and $\text{OPFSEM}_\text{Imst}$ \citep{papa2012efficient}. Two plant datasets, soybean leaf diseases (SOYBEAN-LEAF, 6 classes with 500 images per class) and soybean herbivorous pests (SOYBEAN-PESTS, a total of 5,000 images for 13 herbivorous pest classes), were collected using UAVs under real field conditions with two different percentages of unlabeled samples, 90\%, and 50\%. Experimental results showed that Inception-V3 with $\text{OPFSEM}_\text{Imst}$ achieved the best performance on the SOYBEAN-LEAF dataset with an accuracy of 98.90\%, compared to an accuracy of 99.02\% obtained by VGG16 \citep{simonyan2014very} with the fully labeled dataset. Similarly, ResNet-50 with TSVM obtained the best accuracy of 90.19\% on the SOYBEAN-PESTS dataset, compared to an accuracy of 93.55\% with ResNet-50 on the fully labeled samples. The results showed that the proposed semi-supervised learning methods have a good generalization ability, especially when the labeled samples are limited.
In \cite{li2021semi}, the pseudo-label approach and few-shot learning \citep{wang2020generalizing} were applied for plant leaf disease recognition with only a few labeled samples and a large number of unlabeled samples. To demonstrate the effectiveness of the proposed approach, 1,000 images per class were randomly selected from PlantVillage \citep{hughes2015open} dataset, a public dataset with 38 classes of plant leaf diseases and healthy crops. The curated dataset was split into a source subset with 28-class labeled samples and a targeted subset with the remaining 10 classes in which only a few samples were labeled (fewer than 20 images). In their work, the authors first pre-trained a CNN-based classifier on the source subset and fine-tuned the model on the target subset through transfer learning \citep{zhuang2020comprehensive}, aiming at recognizing unseen samples. An adaptive selection method was proposed to select unlabeled samples that had prediction confidence higher than 99.5\% and feed them to the pre-trained classifier to obtain the pseudo labels. Then the original labeled samples and pseudo-labeled samples were both fed to fine-tune the model. The proposed approach yielded average accuracies of 90\% and 92.6\%, respectively, at the 5-shot and 10-shot settings, which outperformed baseline methods that only gave an accuracy of 90\% at 80-shot.

\begin{figure*}[!ht]
  \centering
\includegraphics[width=0.95\textwidth]{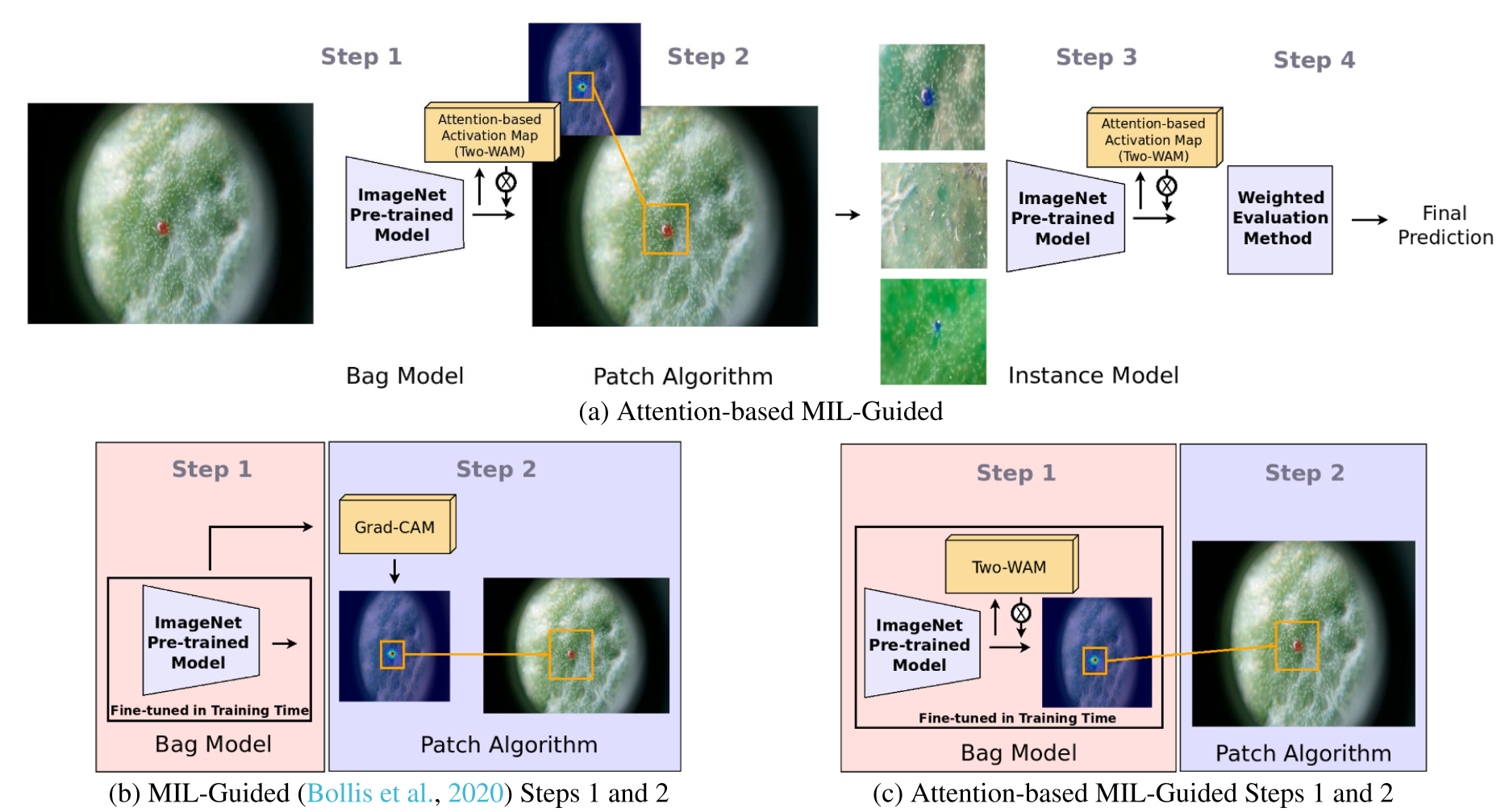}
  \caption{Steps for the Attention-based Multiple Instance Learning Guided approach for citrus mite and insect pest classification \citep{bollis2022weakly}.}
  \label{fig:figure}
  \vspace{-10pt}
\end{figure*}

\textit{\textbf{Weakly-supervised learning. }} 
Multi-instance learning (MIL, Section~\ref{sec:wsl}) has also been evaluated to reduce labeling efforts in plant health detection.
In \cite{lu2017field}, an automatic wheat disease identification and localization system was developed based on the MIL framework with only image-level annotations. To be specific, the input images were first processed by two modified VGG16 \citep{simonyan2014very} models with the fully convolutional network (FCNs) \citep{long2015fully} to generate some spatial score maps, which were then utilized to obtain the disease identification results based on the MIL framework. To localize the disease, a bounding box approximation (BBA) algorithm was developed and implemented using OpenCV\footnote{OpenCV: an open-source computer vision library at \url{https://opencv.org/}.} to obtain the bounding box information. An in-field disease dataset (i.e., Wheat Disease Database 2017 (WDD2017)), consisting of 9,230 images with six common wheat disease classes and a healthy wheat class, was exploited to evaluate the effectiveness of the proposed framework, reporting the mean recognition accuracy up to 97.95\%, which significantly outperformed two conventional VGG16 \citep{simonyan2014very} model variants (93.27\% and 73.00\%, respectively).
In \cite{bollis2020weakly}, the authors applied the MIL framework to automatically detect regions of interest (ROIs) to identify plant disease symptoms. As shown in Fig.~\ref{fig:figure} (b), firstly, pre-trained CNN models (e.g., Inception-v4 \citep{szegedy2017inception}, ResNet-50 \citep{he2016deep}, and MobileNet-v2 \citep{sandler2018mobilenetv2}) were trained on the annotated dataset (original images with image-level labels), resulting in a Bag model, which was applied to generate activation maps for each input image. Image patches (i.e., instances) were extracted based on the Grad-CAM  algorithm \citep{selvaraju2017grad})  to train a CNN model in a fully supervised way. Lastly, a novel weighted evaluation method was proposed to obtain the image class based on the instance probabilities. The proposed framework was evaluated on a new Citrus Pest Benchmark (CPB, including 7,966 mite images of six mite species and 3,455 negative images) and the IP102 database \citep{wu2019ip102} that consists of 75,222 images of 102 classes), yielding an improvement of 5.8\% (from 86.0\% to 91.8\%) for classifying patch instances as compared to the manually-annotated method. 
To detect salient insects of tiny regions, \citep{bollis2020weakly} was extended to an attention-based deep MIL framework in \cite{bollis2022weakly} (Fig.~\ref{fig:figure} (a) and (c)) with only image-level labels. In the new framework, CNN equipped with the novel attention-based activation map architecture (Two-Weighted Activation Mapping (Two-WAM) scheme) were able to dynamically focus their attention only on certain parts of the input images that effectively affect the task \citep{chaudhari2021attentive}. \cite{bollis2022weakly} reported an improvement of at least 16.0\% on IP102 and CPB databases compared to the literature baselines.

In \cite{wu2019crop}, a two-step strategy was proposed for the plant organ instance segmentation and disease identification based weakly supervised approach with only bounding-box labels. In the first stage, GrabCut \citep{rother2004grabcut}, a foreground segmentation and extraction algorithm, was applied to obtain the pixel-level labels based on the annotated bounding boxes. Then, Mask R-CNN \citep{he2017mask} was trained on these labeled samples for organ instance segmentation. With the segmented instances, a lightweight CNN model was subsequently trained to identify the leaf diseases. Applying the proposed framework on a tomato disease dataset, consisting of 37,509 images of ten common disease classes and a healthy class, showed a segmentation accuracy of up to 97.5\% and a recognition accuracy of 98.61\%. However, GrabCut algorithm may fail  if the background is complex or the background and the object are very similar \citep{li2018grab}.

In \cite{kim2020machine}, the weakly supervised learning approach, class activation map (CAM) \citep{zhou2016learning}, was employed to classify and localize online onion disease symptoms only with image-level annotation in a real-time field monitoring system equipped with a high-resolution camera. Through a local wireless Ethernet communication network, the captured onion images were transmitted and then processed, resulting in a dataset of 12,813 cropped images at $224 \times 224$ resolution of six classes (normal growth, disease symptom, lawn, worker, sign, and ground). The weakly supervised learning framework (\citep{zhou2016learning}, Section~\ref{sec:wsl}) took the average values of the final feature maps through the global average pooling (GAP) and trained the classifier to know the importance of each feature map. Four modified VGG16 \citep{simonyan2014very} networks with different network settings were tested with the framework, achieving identification accuracies $mAP@0.5$ ranging from 74.1\% to 87.2\%. Despite the decent performance, CAM \citep{zhou2016learning} algorithm can only be applied to particular kinds of CNN architectures and CNN models like
VGG \citep{simonyan2014very}, which greatly hinders the development of more robust and generalized algorithms.
\vspace{10pt}\\
\underline{\textbf{No supervision}} \vspace{5pt} \\
Self-supervised learning has also been evaluated in plant disease applications, while the conventional unsupervised representation learning approaches (Section~\ref{sec:selfsl}) remain unexplored by the community for plant health applications.

\textit{\textbf{Self-supervised learning. }}
Contrastive learning based on the Siamese network (Section~\ref{sec:selfsl}) has been adopted to reduce label costs for enhanced plant disease recognition. 
In \cite{fang2021self}, a novel self-supervised learning algorithm, cross iterative under-clustering algorithm (CIKICS), was employed for grouping (i.e., clustering) unlabeled plant disease images to save expensive annotation time. Specifically, a batch of feature vectors were first extracted from the input images with ResNet-50 \citep{he2016deep}, and further dimensional reduction was performed with the t-SNE algorithm. Then, a Kernel k-means \citep{dhillon2004unified} algorithm was adopted to cluster unlabeled data into the normal clusters or the abnormal cluster through the CIKICS, in which the normal clusters were pseudo-labeled as the training set, and the abnormal cluster was considered as the predicting set. A CNN-based image classification model was trained on the training set and predicts the clusters for images in the abnormal set. Two more similarity measurements (i.e., similarity score calculated with the CNN-extracted feature space and Siamese network \citep{chicco2021siamese}) were adopted to further improve the accuracy for classifying images in the abnormal cluster. Experimental results on the PlantVillage \citep{hughes2015open} and Citrus Disease Dataset (CDD) \citep{rauf2019citrus} datasets showed that the proposed framework achieved comparable or higher performance than other clustering methods, representing average accuracies of 89.1\%, 92.8\%, and 77.9\%, respectively. However, the training process is not end-to-end and the performance highly depends on the effectiveness of each separate component. 
In \cite{monowar2022self}, an end-to-end deep Siamese model based on the AutoEmbedder \citep{ohi2020autoembedder} was proposed to cluster leaf disease images without manual-labeled annotations. It was trained to distinguish the similarity between the image pairs with high correlation or uncorrelation in a self-supervised way until the model learns class discriminative features, which were then used to generate clusterable feature embeddings that were clustered by the k-means  algorithm \citep{hartigan1979algorithm}. Evaluated on the CDD  dataset \citep{rauf2019citrus}, the proposed approach achieved a clustering accuracy of 85.1\%, outperforming other state-of-the-art self-supervised approaches,including CIKICS \citep{fang2021self} (13.9\%) and SimSiam \citep{chen2021exploring} (58.2\%).

In \cite{kim2022instance}, a novel self-supervised plant disease detector was proposed by leveraging conditional normalizing flows \citep{kobyzev2020normalizing}. Instead of inputting raw images into the flow model, a CNN model (i.e., Wide-ResNet-50-2)  was employed to extract the multi-scale features from the images, which was pre-trained on the ImageNet \citep{deng2009imagenet} dataset using the Simsiam (\citep{chen2021exploring}, Section~\ref{sec:selfsl}) algorithm. The flow model was trained to learn to map complex distributions of image features to simple likelihoods, which indicated whether the input images were infected or healthy. The proposed approach was evaluated on the BRACOL \citep{esgario2020deep} and PlantVillage \citep{hughes2015open} datasets, yielding improvements of detection accuracies by 1.01\% to 14.3\% as compared to the benchmark without self-supervised pre-training.



\subsubsection{Weed and crop management}
Weeds can significantly reduce crop production as they compete for crucial resources like water and nutrients, and may serve as hosts for pests and diseases \citep{chen2022performance, coleman2019using, colemanweed}. To address this issue, machine vision-based weed control is emerging as a promising solution, allowing for accurate identification and localization of weed plants and site-specific, individualized treatments such as spot spraying or high-flame laser weed killing. However, the development of robust machine vision systems is heavily reliant on large volumes of labeled image datasets \citep{westwood2018weed, chen2022performance, dang2023yoloweeds}, which is often cost-expensive and time-consuming. As such, there is a growing research interest in developing label-efficient learning algorithms for weed (crop) recognition.
\vspace{10pt} \\
\underline{\textbf{Weak supervision}} \vspace{5pt} 
\\
Algorithms with weak supervision, such as active learning and semi-supervised learning approaches have been widely explored. On the other hand,  weakly-supervised learning  (Section~\ref{sec:wsl}) for weed and crop recognition remains largely unexplored.
\\
\textit{\textbf{Active learning. }} 
To reduce the expensive labeling costs, in \cite{yang2022dissimilarity}, the \underline{d}issimilarity-\underline{b}ased \underline{a}ctive \underline{l}earning (DBAL, Section~\ref{sec:al}) framework was applied for weed classification, which only required a small amount of representative samples to be selected and labeled. Specifically, pre-trained CNN models such as VGG \citep{simonyan2014very} and ResNet \citep{he2016deep} were first employed through transfer learning \citep{zhuang2020comprehensive, liu2022yolov5} to extract the feature representation from both labeled and unlabeled samples. A binary classifier was then trained on the extracted features to distinguish which group the features come from. Then, the top-k most representative samples were selected through the calculation of the Euclidean distance between the cluster centroid (generated by the k-means \citep{hartigan1979algorithm} algorithm) of the labeled samples and the unlabeled samples. These selected samples were then labeled and used to re-train the CNN models. The process continued until certain accuracy was achieved or the maximum iteration number was reached. The proposed approach was able to achieve classification accuracies of 90.75\% and 98.97\%, respectively, on DeepWeeds \citep{olsen2019deepweeds} dataset and Grass-Broadleaf dataset\footnote{Grass-Broadleaf dataset: \url{https://www.kaggle.com/datasets/fpeccia/weed-detection-in-soybean-crops}} with only 32\% and 27.8\% labeled samples, which compared favorably with the results obtained by training using fully labeled datasets with classification accuracies of 91.5\% and 99.52\%, respectively. Despite promising results, the usage of k-means \citep{hartigan1979algorithm} algorithm was sensitive to the initial choice of centroids and may not perform well on datasets with complex structures \citep{arthur2007k}.


\textit{\textbf{Semi-supervised learning. }} 
In \cite{perez2015semi}, a semi-supervised framework was developed for weed mapping and crop row detection in sunflower crops with UAV-acquired images based on improved Hough transform and Otsu’s method \citep{otsu1979threshold}. 
Different machine learning algorithms and training strategies, including supervised k-nearest and SVM, semi-supervised linear SVM, and unsupervised K-means \citep{hartigan1979algorithm} algorithms, were investigated and compared for classifying the pixels into crops, soils, and weeds. The semi-supervised approach yielded the best average mean average error (MAE) of 0.1268 as compared to 0.1697, 0.1854, and 0.1962 of supervised k-nearest, SVM, and unsupervised learning (e.g., k-means), respectively. However, the paper did not investigate the effects of different proportions of labeled and unlabeled samples on the developed semi-supervised approach.

The teacher-student framework (Section~\ref{sec:semi}) was also explored for weed detection. In \cite{hu2021powerful}, the authors combined the image synthesis and semi-supervised learning framework for site-specific weed detection without manually labeled images. A novel cut-and-paste image synthesis approach was proposed to generate high-fidelity plant images with target backgrounds and labels. The noisy teacher-student framework \citep{xie2020self} was then adopted to train the Faster-RCNN \citep{ren2015faster} for semi-supervised weed detection. More specifically, a teacher model was first trained on the synthetic weed images and used to generate pseudo-bounding box labels for the unlabeled images. A student model was then initialized with the teacher's model weights and then jointly trained on the synthetic and pseudo-labeled images. The teacher model was also updated with the student model during the training process and was then used to update the pseudo labels. The above process repeated until satisfying performance was achieved. Experimental results on a self-collected weed dataset showed that the proposed semi-supervised approach achieved a mAP of 46.0\% with only synthetic images, which was comparable to the supervised model trained using the fully-labeled real weed dataset with only an mAP of 50.9\%. However, the synthetic images were in low resolution and poor quality, which may limit the effects of the developed detection algorithm. Recent advanced data generation approaches, such as GANs \cite{lu2022generative, xu2023comprehensive} and diffusion models \citep{chen2022deep} may be a promising way for high-fidelity image generation and improve this semi-supervised weed detection framework.

In \cite{nong2022semi}, a semi-supervised semantic segmentation algorithm, SemiWeedNet, was developed for pixel-wise crop and weed segmentation by utilizing a large amount of unlabeled data. The state-of-the-art image segmentation framework, DeepLabv3+ \citep{chen2018encoder}, was applied to encode both labeled and unlabeled images, which were then incorporated into the cross-entropy loss and consistency regularization loss \citep{chen2020simple}, respectively. A joint optimization loss was then proposed to build with the two losses to achieve a balance between labeled and unlabeled samples. In addition, an online hard example mining (OHEM) strategy was proposed to prioritize the hard samples. Tested on a public dataset WeedMap \citep{sa2018weedmap}, which contains 289 pixel-wise labeled UAV images, SemiWeedNet achieved a mean Intersection-over-Union (mIoU) of 69.2\% with only 20\% of labeled samples, close to the performance of the fully supervised baseline (i.e, an mIoU of 70.0\%). \vspace{10pt} \\
\underline{\textbf{No supervision}} \vspace{5pt} \\
Unsupervised learning: unsupervised representation learning and self-supervised learning are both evaluated for weed and crop recognition.
\\
\textit{\textbf{Unsupervised representation learning. }}
In \cite{bah2018deep}, an unsupervised clustering algorithm was developed for automatic inter-row weed detection in the bean and spinach fields with UAV-acquired images. Specifically, the simple linear iterative clustering (SLIC) algorithm \citep{achanta2012slic} was adopted to delimit the crop rows and generate crop masks with the crop lines produced by a normalized Hough transform. Then, the inter-weeds were determined in the regions that do not intersect with the crop masks. A dataset was constructed by the segmented inter-row weeds and crops and used to train a binary CNN classifier (i.e., ResNet-18 \citep{he2016deep}). Experimental results showed that the proposed approach obtained comparable performance to the supervised learning baseline, with small accuracy differences of 1.5\% and 6\% in the spinach and bean fields, respectively. Despite impressive results in detecting inter-row weeds, the approach was highly dependent on parameter tuning for conventional computer vision algorithms, such as the Hough transform used for crop and weed segmentation.

An unsupervised weed distribution and density estimation algorithm was proposed in \citep{shorewala2021weed} without pixel-level annotations as conventional image segmentation algorithms. The unsupervised clustering algorithm \citep{kanezaki2018unsupervised} was employed to cluster the pixels of the input images into two classes (foreground (i.e., crops and weeds) and background (e.g., soil and other non-vegetation pixels)). The extracted vegetation pixels were divided into small non-overlapped tiles that were used for training ResNet-50 \citep{he2016deep} to further classify them as crops or weeds. The weed density estimations were then computed with the recognized weed pixels. The proposed approach was validated on two datasets, the Crop/Weed Field Image dataset \citep{haug2014crop} and the Sugar Beets dataset \citep{chebrolu2017agricultural}, and achieved a maximum weed density estimation accuracy of 82.13\%. Despite the decent performance being achieved, the performance was largely affected by the size of the tiles.

\textit{\textbf{Self-supervised learning. }}
In \cite{guldenring2021self}, a self-supervised contrastive learning framework, SwAV \citep{caron2020unsupervised} (Section~\ref{sec:selfsl}), was explored for  plant classification and segmentation. Two backbone networks, i.e., ResNet-34 and xResNet-34, were evaluated and initially pre-trained on the ImageNet \citep{deng2009imagenet} (i.e., pre-text tasks) in a self-supervised way and then fine-tuned on the agricultural datasets (i.e., downstream tasks). Evaluated on three agricultural datasets, DeepWeeds \citep{olsen2019deepweeds}, Aerial Farmland \citep{chiu2020agriculture}, and a self-collected Grassland Europe dataset with online images, \cite{guldenring2021self} yielded the best Top-1 accuracies of 94.9\%, 70.6\%, and 86.4\%, which were higher than that training without ImageNet pre-training (accuracies of 94.4\%, 68.4\%, and 85.5\%, respectively). Additionally, the authors showed that the proposed framework was also effective in semi-supervised settings, by using limited labeled samples to fine-tune the networks in a supervised way. They concluded that only with 30\% labeled data of DeepWeeds, the pre-trained SwAV was able to outperform the classical transfer learning and fully supervised approaches.

To demonstrate the performance of unsupervised clustering algorithms on the weed classification, in \cite{dos2019unsupervised}, two CNN-based unsupervised learning algorithms, JULE \citep{yang2016joint} and DeepCluster \citep{caron2018deep} were benchmarked and evaluated on two public datasets, Grass-Broadleaf dataset \citep{dos2017weed} and DeepWeeds dataset \citep{olsen2019deepweeds}. In general, DeepCluster showed better performance than the JULE method. For example, DeepCluster achieved an accuracy of 83.4\% and normalized mutual information (NMI) of 0.43 on the Grass-Broadleaf dataset, compared to JULE with an accuracy of 63.5\% and NMI of 0.28. The authors also proposed a semi-automatic data labeling framework based on the clustered data to reduce the cost of manual labeling for weed discrimination. Specifically, DeepCluster was first used to group images into  clusters, and the representative samples in each cluster were then labeled by a human expert. Experimental results showed that hundreds of speedup on data labeling was achieved by setting the number of clusters to be much smaller than the number of samples in the dataset.

In \cite{marszalek2022self}, SimSiam \citep{chen2021exploring} (Section~\ref{sec:selfsl}) was employed for domain adaptation in crop classification. The network with a transformer-like encoder network was trained on data from previous years and made predictions on the later years, thus expensive labeling costs for new years were saved. The yield and climatological dataset \citep{marszalek2022prediction} with various climatological conditions for the years 2016, 2017, and 2018 was used to validate the proposed approach, showing that the proposed approach trained on data from 2016 and 2017 with suitable augmentation techniques was able to achieve an overall accuracy of 71\% on data from the year 2018. As large as 16\% further improvement is achieved when the model was fine-tuned on 5\% labeled data from the year 2018, very close to the accuracy of 93\% achieved by training on the entire data (years 2016-2018).

\begin{figure*}[!ht]
  \centering
  \includegraphics[width=0.75\textwidth]{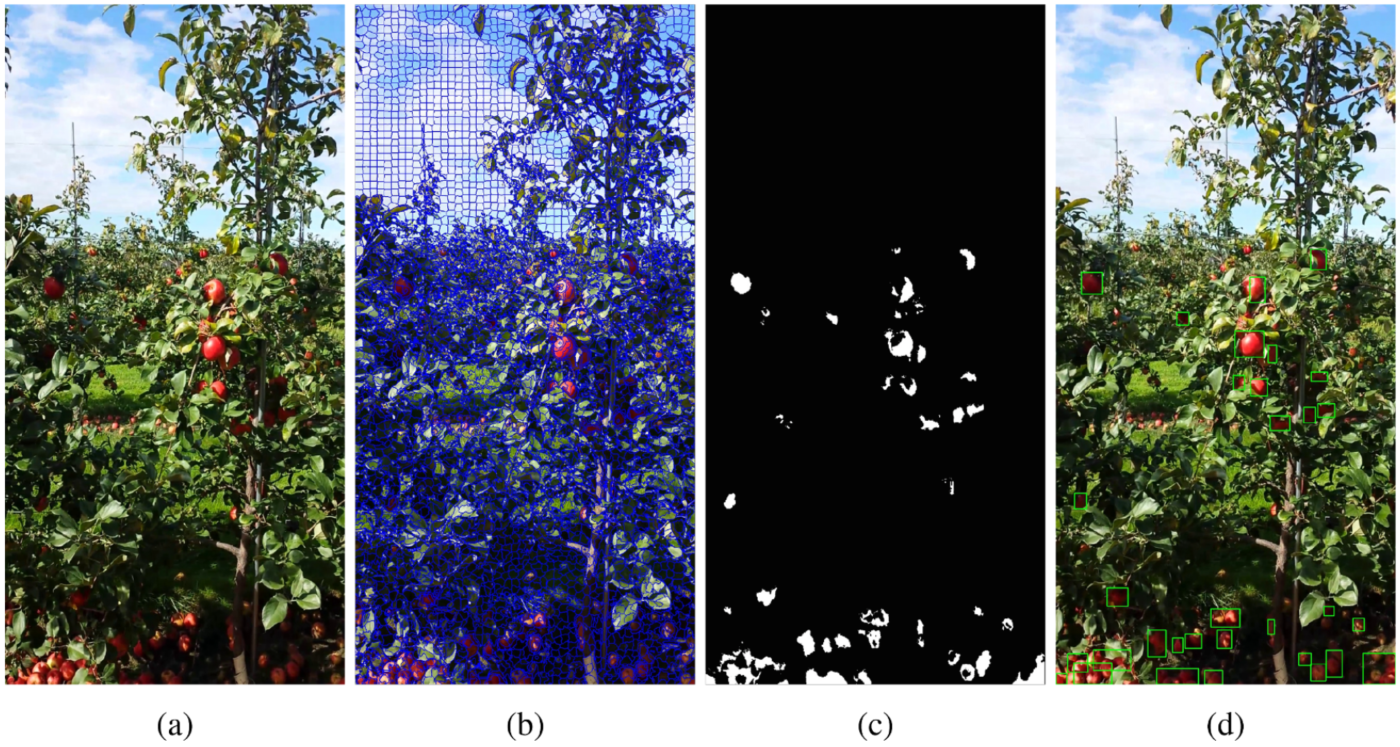}
  \caption{The pipeline of the semi/unsupervised learning algorithm for estimating apple counts for yield estimation \citep{roy2019vision}.}
  \label{fig:figure2}
  \vspace{-10pt}
\end{figure*}

\subsubsection{Fruit detection}
In-orchard fruit detection is an important yet challenging task for automated fruit harvesting due to the unstructured nature of the orchard environment and variations in field lighting conditions \citep{zhang2022algorithm}. In recent years, deep learning-based object detectors and segmentation networks have been extensively studied for fruit detection, particularly in the context of robotic harvesting, fruit counting, and yield estimation \citep{koirala2019deep,maheswari2021intelligent, chu2023o2rnet}. However, labeling large datasets for training these models can be expensive and time-consuming. To address this issue, researchers have turned to label-efficient learning algorithms that utilize weak supervision signals (see Fig.~\ref{fig:app1}). These approaches can achieve satisfactory performance while reducing the need for manual effort in data collection and labeling.

\textit{\textbf{Semi-supervised learning. }}
In \cite{roy2019vision}, a semi-supervised clustering framework was developed to recognize apples for yield estimation with video clips acquired in natural apple orchards under various color  and lighting conditions. As shown in Fig.~\ref{fig:figure2}, the input image (Fig.~\ref{fig:figure2} (a)) was firstly over-segmented into SLIC superpixels \citep{achanta2012slic} (Fig.~\ref{fig:figure2} (b)) using the LAB colorspace. The superpixels were then modeled as a Gaussian Mixture Model (GMM) \citep{bilmes1998gentle} and clustered into similar colored superpixels, representing different semantic components such as apples, leaves, branches, etc. The first few frames of the video clips were manual-labeled at pixel level and used to classify and localize apple pixels using GMM \citep{bilmes1998gentle}, expectation maximization (EM) \citep{bilmes1998gentle}, and heuristic Minimum Description Length (MDL) \citep{grunwald2005minimum} algorithms (Fig.~\ref{fig:figure2} (c-d)), and the apple counts across video frames were merged considering the camera motion captured by matching SIFT features \citep{lowe1999object}. The proposed approach was trained on a self-collected video dataset containing 76 apple trees with a mixture of green and red apples, resulting in four different manually annotated datasets with varied apple colors and geometry structures, showing that it can achieve counting accuracies ranging from 91.98\% to 94.81\% on different datasets. 

In \cite{casado2022semi}, three semi-supervised learning approaches, i.e., PseudoLabeling \citep{lee2013pseudo}, Distillation \citep{hinton2015distilling} and Model Distillation \citep{bucilamodel}, and 13 CNN architectures (e.g., DeepLabV3+ \citep{chen2018encoder}, HRNet \citep{sun2019high}, and U-Net \citep{ronneberger2015u}) were evaluated for the object segmentation (i.e., bunches, poles, wood, leaves, and background) with natural images obtained in a commercial vineyard. A grape dataset was collected and open-sourced, containing 405 natural color images with 85 manually annotated and 320  unlabelled. Compared to training only with labeled samples, the semi-supervised learning approaches can improve the mean segmentation accuracy MSA by at least 5.62\% with the usage of a large number of unlabeled samples. DeepLabV3+ with Efficientnet-B3 backbone trained with Model Distillation \citep{bucilamodel} yielded the highest  MSA of 85.86\% on the bunch/leave segmentation tasks. HRNet obtained the highest MSA of 85.91\% for the object segmentation tasks of all classes. However, the performance of the semi-supervised learning approaches on the all-class segmentation task was not reported.

In \cite{khaki2021deepcorn}, 
the noisy student training algorithm \citep{xie2020self} (Section~\ref{sec:semi}) was employed for on-ear corn kernel counting and yield estimation. The counting results were obtained with the Euclidean loss between the estimated density map (generated by a lightweight VGG-16 network \citep{simonyan2014very}) and ground truth. A corn kernel dataset, containing 154,169 corn kernel images, was collected where 30,000 of them were labeled with ground truth density maps using \citep{boominathan2016crowdnet}. The noisy student-teacher framework was employed to generate pseudo-density maps. Experimental results showed that the proposed approach yielded the lowest mean absolute error (MAE) and root mean squared error (RMSE) of 41.36 and 60.27, outperforming the case of training with only the labeled data with MAE and RMSE of 44.91 and 65.92, respectively. However, the developed approach still relied heavily on a large number of labeled images (30,000 images), which can be time-consuming and labor-intensive, especially when dealing with large and complex datasets.

\textit{\textbf{Weakly-supervised learning. }}
In \cite{bellocchio2019weakly}, a weakly supervised fruit counting framework was developed with only image-level annotations. The input image was processed by a three-branch counting CNN across three different image scales, i.e., 1, $\frac{1}{2}$, and $\frac{1}{4}$. The counting results were set to be consistent at all the image scales. To further ensure counting consistency, an image-level binary classification model was trained on the image-level annotations. The output of the binary classifier was supposed to be equal to the binarized counting results of the multi-branch counting CNN constrained with a consistency loss function. Experimental results showed that the proposed approach achieved comparable counting accuracy as compared to two supervised approaches while significantly outperformed a weakly supervised approach on three fruit datasets (i.e., apples, almonds, and olives).

The work \cite{bellocchio2019weakly} was then extended for unseen fruit counting in \cite{bellocchio2020combining} with weak supervision and unsupervised style transfer methods on  image-level only annotations. Based on \cite{bellocchio2019weakly}, a Peak Stimulation Layer (PSL) \citep{zhou2018weakly} was adopted to facilitate the model training. To adopt the trained model for different fruit species, the CycleGAN \citep{zhu2017unpaired} was employed for the unsupervised domain adaptation,  transferring the known fruit species to unseen fruit species. It was trained with a designed presence-absence classifier (PAC) that discriminates images containing fruits or not. The proposed approach was validated on four different datasets: two almond datasets, an olive dataset, and an apple dataset, showing a superior counting accuracy than their previous approach \citep{bellocchio2019weakly}.

In \cite{bellocchio2022novel}, a weakly supervised learning framework, \underline{W}eakly-\underline{S}upervised \underline{F}ruit \underline{F}ocalization \underline{N}etwork (WS-FLNet), was developed for automatic fruit detection, localization, and yield estimation. Specifically, U-Net \citep{ronneberger2015u} was first employed to generate pixel-wise activation maps that implicitly indicated the locations of fruits trained using binary cross-entropy (BCE) loss with only image-level annotations.
The developed framework was validated on ACFR-Mangoes dataset \citep{stein2016image} and ISAR-Almonds \citep{bellocchio2019weakly} dataset. On ACFR-Mangoes dataset, the proposed approach presented relatively lower performance (with a MAE, a RMSE, and a MAEP of 58.61\%, 69.79\%, and 33.80\%, respectively) compared to the supervised approach (i.e., with a MAE, a RMSE and a MAEP of 27.66\%, 15.52\%, and 35.16\%, respectively) due to grouped and partially overlapped fruits. However, on the ISAR-Almonds dataset, the authors showed that the proposed approach (with a MAE, a RMSE, and a MAEP of 71.83\%, 63.68\%, and 81.35\%, respectively) outperformed their previous work \citep{bellocchio2019weakly} (with a MAE, a RMSE and a MAEP of 88.33\%, 83.29\%, and 114.05\%, respectively) by a large margin.

\begin{figure*}[!ht]
  \centering
\includegraphics[width=0.95\textwidth]{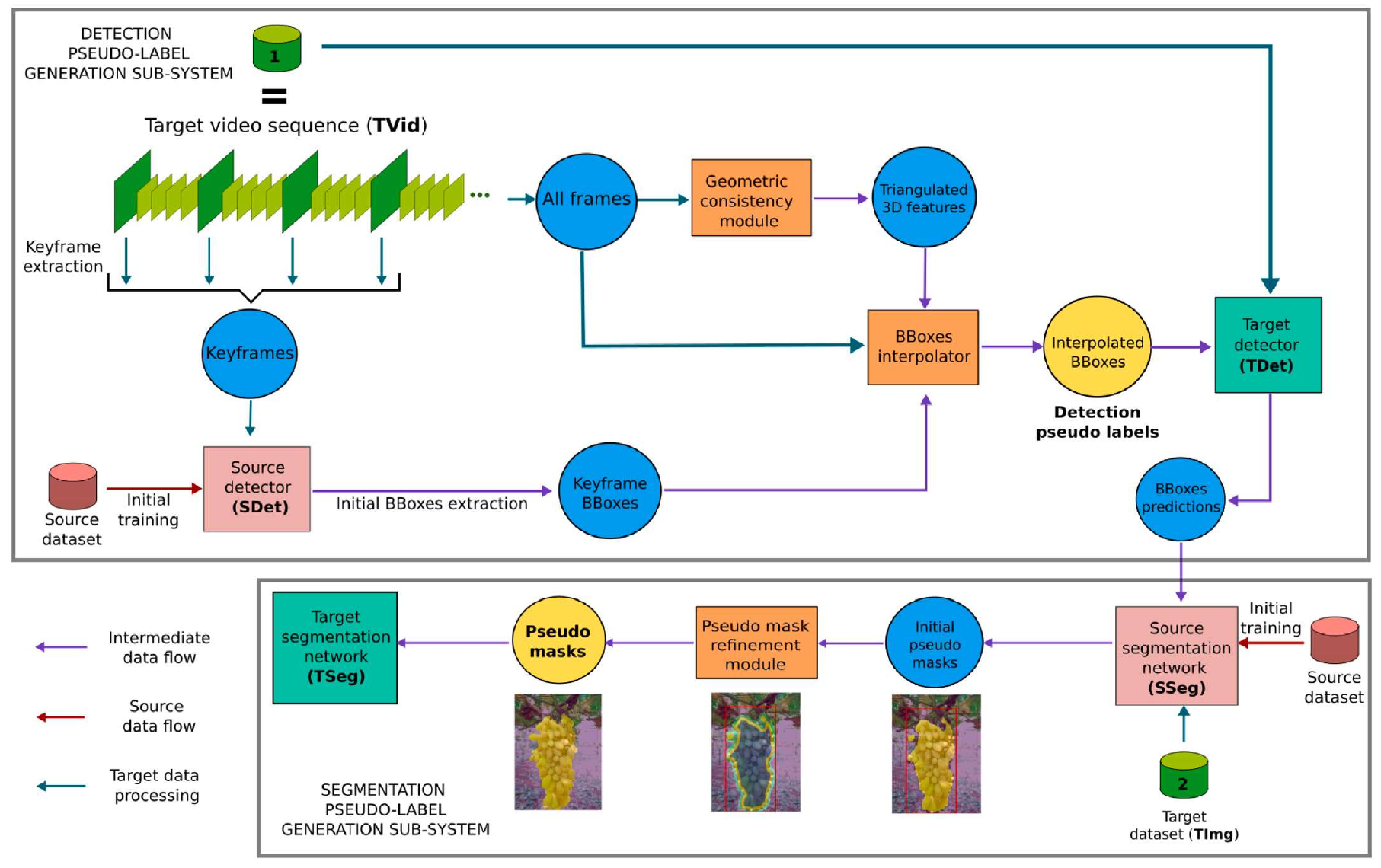}
  \caption{The framework of a weakly-supervised learning algorithm for detecting, segmenting, and tracking table grapes \citep{ciarfuglia2023weakly}.}
  \label{fig:figure8}
  \vspace{-10pt}
\end{figure*}

In \cite{ciarfuglia2023weakly}, a weakly-supervised learning framework was proposed to detect, segment, and track table grapes. As shown in Fig.~\ref{fig:figure8}, YOLOv5s \citep{jocher2020yolov5} was applied for the detection tasks and trained on a small amount of labeled data from a similar dataset, WGISD \citep{santos2020grape}. Then, the trained YOLOv5s model was used to create pseudo labels for two target datasets: a video dataset containing 1469 frames and an image dataset with 134 images  labeled with bounding boxes and 70 images labeled at the pixel level. To accurately associate grape instances across different video frames, two algorithms, SfM \citep{santos2020grape} and DeepSORT \citep{wojke2017simple}, were utilized to interpolate the bounding boxes for the remaining video frames. For accurate yield estimation, Mask R-CNN \citep{he2017mask} was adopted and trained on the source dataset and generated pseudo masks for the target datasets. Three pseudo mask refinement strategies, i.e., morphological dilation, SLIC \citep{achanta2012slic}, and GrubCut \citep{rother2004grabcut}, were then evaluated to refine the segmentation masks based on the bounding boxes obtained in the detection stage.
Experiments conducted on the image and video datasets showed that using pseudo-labels, the ${mAP}_{0.5}$ was increased from 69\% to 77.0\% and from 55.0\% to 65\%, respectively, compared to training only on the source dataset. For the tracking performance across the video frames, the SfM algorithm \citep{santos2020grape} using the pseudo-labels achieved the lowest tracking error of 9.0\%, compared to the baseline approach trained without the pseudo-labels with an error of 38.0\%. For the segmentation performance, the refinement trick GrabCut yielded the highest improvement of $mAP@[0.5:0.95]$ by 1.13\% and $mAP@0.75$ by 4.58\%. Although good performance is achieved, the developed approach is not end-to-end and the final performance  highly depends on the performance of each individual component.

In \cite{bhattarai2022weakly}, the authors proposed a weakly-supervised learning framework for flower and fruit counting in highly unstructured orchard environments with only image-level annotations. Instead of counting by object detection \citep{farjon2020detection} with dense bounding box annotations, a regression-based CNN network based on VGG16 \citep{simonyan2014very} was proposed to estimate the count for a whole image without inferring explicit information on the location of the objects based on Score-CAM \citep{wang2020score} (Section~\ref{sec:wsl}) and Guided Backpropagation \citep{springenberg2014striving}. Experimental results on self-collected apple flower and fruit canopy image datasets showed that the proposed approach was able to learn the underlying image features corresponding to the apple flower or fruit locations and achieves the lowest MAE of 12.0 and 2.9 on the flower and fruit datasets, respectively.


\subsubsection{Aquaculture}
Precision aquaculture farming presents a unique set of challenges when it comes to using imaging technology for detecting and monitoring aquatic species \citep{li2021automatic, fore2018precision}. Underwater conditions can be adverse, with poor illumination and low visibility in turbid water, as well as cluttered backgrounds, making it difficult to acquire high-fidelity and high-contrast images. Furthermore, the scarcity of labeled aquaculture images available adds to the complexity of underwater species recognition tasks. To address these challenges, researchers have turned to active learning algorithms that can achieve accurate and reliable species recognition with a smaller number of labeled images \citep{li2021automatic, kong2022recurrent}.

\begin{figure*}[!ht]
  \centering
  \includegraphics[width=0.75\textwidth]{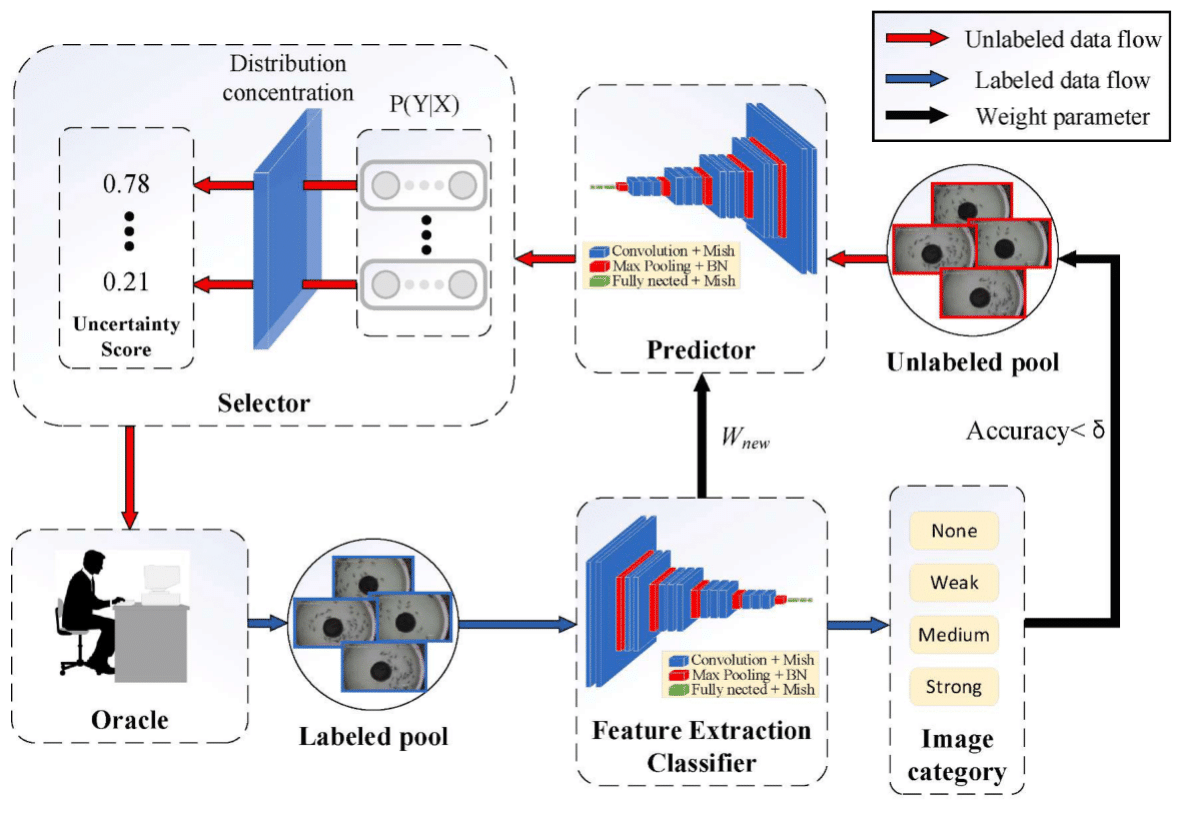}
  \caption{The framework of an active learning-based algorithm for fish feeding status classification \citep{kong2022recurrent}.}
  \label{fig:kong2022recurrent}
  \vspace{-10pt}
\end{figure*}

\textit{\textbf{Active learning. }}
In \cite{kong2022recurrent}, an active learning framework was proposed to classify fish feeding status for sustainable aquaculture \citep{fore2018precision}. The objective was to train a CNN model to classify the input images into four categories: no feeding, weak feeding, medium feeding, and strong feeding by utilizing a large amount of unlabeled data. To collect the image dataset, 50 fish (i.e., oplegnathus punctatus) were kept in a well-controlled tank with a high-resolution camera mounted on the top. Overall, 3,000 image samples were collected, among which 100 were labeled and placed into the labeled pool whereas the remaining 2,900 images were placed into the unlabeled pool. As shown in Fig.~\ref{fig:kong2022recurrent}, the proposed approach consisted of two major components: a predictor and a selector. The predictor was made of a CNN-based prediction network (i.e., VGG16 \citep{simonyan2014very}), which was trained on a small number of labeled samples and tested on the test subset. If the test accuracy was smaller than a predefined threshold $\delta$, then the images in the unlabeled pool were fed into the selector to obtain the uncertainty scores of each sample. Then, the most representative (highly uncertain) unlabeled samples will be selected and labeled by an oracle and added to the labeled pool. The predictor was trained with the newly labeled pool and the process repeats until the accuracy threshold on the test subset was satisfied. Experimental results showed that the proposed algorithm was able to achieve a classification accuracy of 98\% with only 10\% of labeled samples, which greatly reduced the labeling costs. 

\begin{table*}[!ht]
  \centering
  \caption{Application of label-efficient learning in precision agriculture.}
  \label{fig:app1}
  \includegraphics[width=0.85\textwidth]{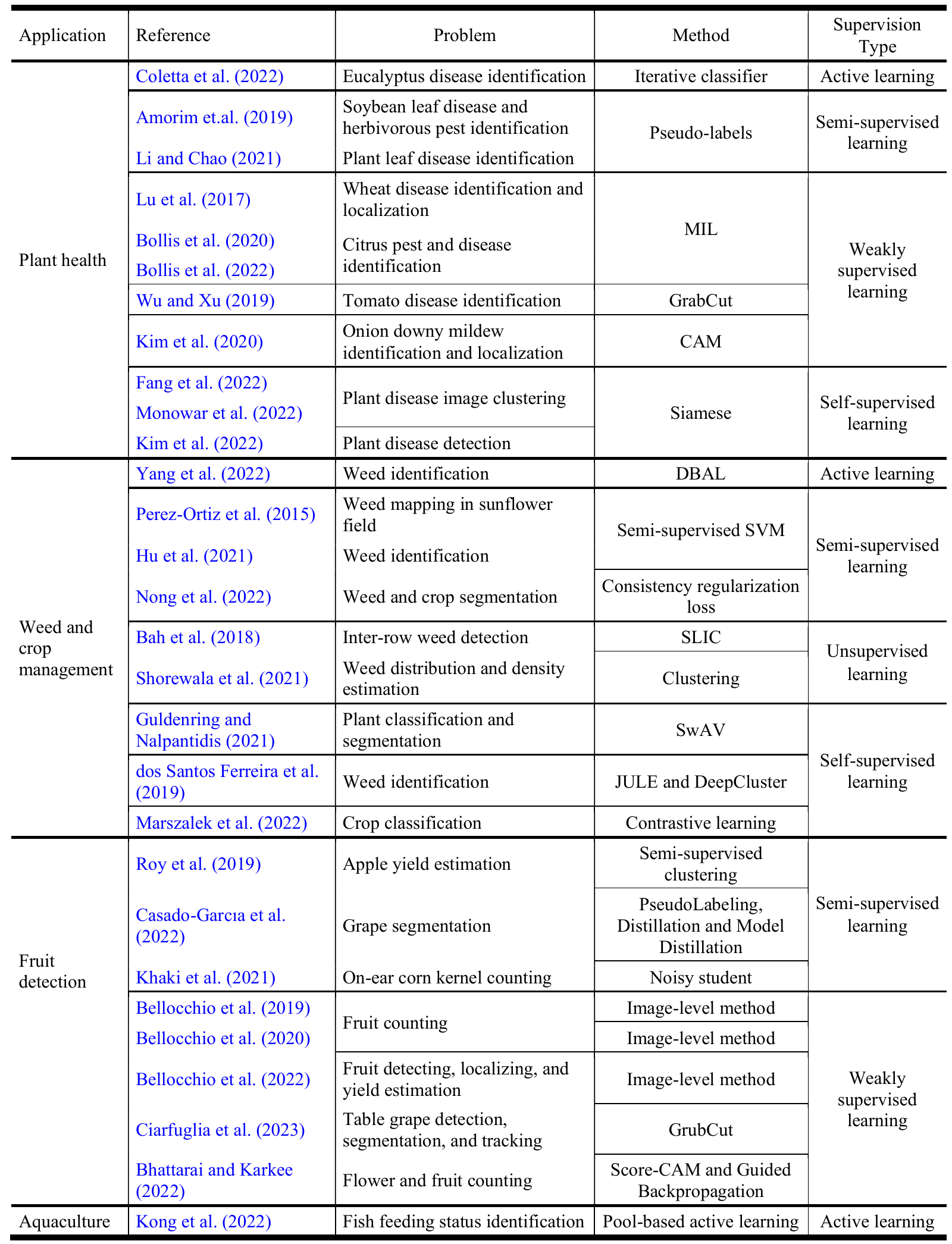}
\end{table*}


\subsection{Plant Phenotyping}
Plant phenotyping/phenomics in crop breeding involves the quantification of various plant phenotypes, such as growth dynamics and stress resistance, resulting from the complex interactions between genetics and environmental conditions. Imaging technologies are critical for achieving high-throughput, automated quantification of plant phenotypes, thereby accelerating breeding processes and bridging the genotype-phenotype gap \citep{minervini2015image, das2019leveraging, furbank2011phenomics}. However, extracting meaningful phenotypic information from images is still challenging due to factors such as lighting variations, plant rotations, and occlusions \citep{das2019leveraging}. To address this, plant scientists have turned to data-driven machine learning methods for effective feature extraction, plant trait identification, and quantification \citep{rawat2022useful}. However, these methods typically require large amounts of labeled training samples, which can be expensive and time-consuming to acquire. To overcome this challenge, label-efficient learning algorithms (Table~\ref{tab:table2}), such as those with weak supervision and no supervision labels, have been proposed and are being actively researched in the plant phenotyping community \citep{rawat2022useful}. \vspace{10pt} \\
\underline{\textbf{Weak supervision}} \vspace{5pt} \\
\textit{\textbf{Active learning. }}
In \cite{rawat2022useful}, four uncertainty-based active learning algorithms were benchmarked and evaluated for plant organ segmentation on three plant datasets, ACFR Orchard Fruit Dataset (Apple) dataset \citep{bargoti2017image}, UTokyo Wheat 2020 (Wheat) dataset \citep{david2021global}, and UTokyo Rice 2013 (Rice) dataset \citep{desai2019automatic}. The approach followed the standard active learning framework as described in Section~\ref{sec:al}. The least confidence method \citep{lewis1995sequential}, margin-based method (MAR) \citep{scheffer2001active}, Shannon’s entropy \citep{shannon2001mathematical}, and deep Bayesian active learning \citep{gal2017deep} approaches were evaluated and used to calculate the informativeness score (IS) of each unlabeled sample. Then, the sample with the maximum IS was selected and labeled by a human expert. Deeplabv3+ \citep{chen2018encoder} with ResNet50 \citep{he2016deep} backbone was employed for segmentation tasks. On the apple and wheat datasets, the MAR-based approach achieved 0.43\% and 0.53\% increases in the intersection over union (IoU) compared to the random sampling method. However, on the rice dataset, random sampling showed better performance than the active learning-based approaches. The authors concluded that, due to imbalanced datasets, there was no clear winner among the active learning methods across the datasets. 

In \cite{chandra2020active}, an active learning approach based on point supervision (Fig.~\ref{fig:annotation_time}) was proposed for cereal panicle detection to reduce the expensive annotation costs. Three uncertainty estimation methods, i.e., max-variance, max-entropy, and max-entropy-variance were explored to estimate the uncertainties of each unlabeled sample, and the most uncertain samples were selected and labeled by an oracle. To further reduce the labeling costs, the authors adopted a weakly supervised approach \citep{papadopoulos2017training} (i.e., point supervision), in which object centers were ground-truthed by points instead of dense bounding box labeling. The labeled samples were used to train a CNN-based object detector, i.e., Faster R-CNN \citep{ren2015faster}. The authors validated their approach on two public datasets, the Wheat dataset \citep{madec2019ear} and the Sorghum dataset \citep{guo2018aerial}. Compared to the baseline method (i.e., 81.36\% mAP and $106.76~h$ annotation time), the proposed methods achieved better performance  with 55\% labeling time-saving at the same time on the Sorghum dataset, i.e., about 86\% mAP and $60~h$ annotation time. On the Wheat dataset, the proposed approach saved up to 50\% labeling time (less than $12.75~h$) while also achieving superior performance compared to the baseline methods (i.e., 73.31\% mAP and $29.14~h$ annotation time). 

In \cite{blok2022active}, an uncertainty-aware active learning method \citep{morrison2019uncertainty} was employed for instance segmentation of broccoli heads, and the software was made publicly available\footnote{Software: \url{https://github.com/pieterblok/maskal}}.
Mask R-CNN \citep{he2017mask} with ResNeXt-101 \citep{xie2017aggregated} as the backbone was adopted for the instance segmentation task. Images with the most uncertainties \citep{morrison2019uncertainty} were sampled and labeled for the model training. The proposed framework was validated on a self-collected dataset, consisting of 16,000 RGB broccoli images of five broccoli classes (including healthy, damaged, matured, broccoli head with cat-eye, and broccoli head with head rot). With only 17.9\% training data, the proposed approach was able to achieve 93.9\% mAP as the fully supervised method. It was able to achieve comparable performance as the random sampling approach with only $1/3$ of samples (900 v.s. 2700).

\textit{\textbf{Semi-supervised learning. }} 
The pseudo-labeling approach \citep{lee2013pseudo} discussed in Section~\ref{sec:semi} is widely explored for plant phenotyping applications.
In \cite{fourati2021wheat}, the semi-supervised pseudo-labeling approach \citep{lee2013pseudo} was employed for wheat head detection. Both one-stage object detector (EfficientDet \citep{tan2020efficientdet}) and two-stage object detector (Faster R-CNN \citep{ren2015faster}) were evaluated on the detection tasks. The pseudo-labeling approach was used to predict the labels for the unlabeled samples in the test set to enrich the training set and re-train the models. Evaluated on the Global Wheat Head Detection (GWHD 2021) \citep{david2020global} dataset, Faster-RCNN and EfficientDet 
yielded, respectively, 1.22\% and 0.79\% improvements in detection accuracy compared to training without semi-supervised learning. The developed framework ranks in the top 6\% in the Wheat Head Detection challenge\footnote{Wheat Head Detection challenge: \url{https://www.kaggle.com/competitions/global-wheat-detection/overview}}. However, in the study, the systematic performance comparison between Faster R-CNN and EfficientDet was not reported.
In \cite{najafian2021semi}, the pseudo-labeling approach \citep{lee2013pseudo} was employed for wheat head detection with video clips and only one video frame of each video clip was labeled with the remaining unlabeled. These unlabeled video frames were then pseudo-labeled with a dedicated two-stage domain adaptation approach. With the developed framework, the one-stage object detector (YOLO \citep{redmon2016you}) obtained a $\text{mAP}@0.5$ of 82.70\% on the GWHD 2021 dataset \citep{david2020global}, which outperformed the baseline method that only gives a $\text{mAP}@0.5$ of 74.10\%.
In \cite{li2022leaf}, the pseudo-labeling approach \citep{lee2013pseudo} was applied for leaf vein segmentation with only a few labeled samples. To enhance segmentation performance, an encoder-decoder network, called Confidence Refining Vein Network (CoRE-Net), was implemented and trained in a two-phased training framework. In the first stage, the model was warm-start trained on a few labeled samples (less than 10 samples for each leaf class) in a supervised manner. Then, the pre-trained model was used to generate pseudo labels for the unlabeled samples. \cite{li2022leaf} collected and released the leaf vein dataset (LVD2021), comprising 5406 images with 36 leaf classes. An improvement of up to 9.38\% accuracy was achieved using the semi-supervised learning approach. 

In \cite{ghosal2019weakly}, an active learning-based weakly supervised learning framework was proposed for sorghum head detection and counting with UAV-acquired images. A CNN model, RetinaNet \citep{lin2017focal} with the feature pyramid network (FPN) \citep{lin2017feature} as the backbone, was first trained on a single image to get a semi-trained model that was used to generate pseudo labels (i.e., bounding boxes) for randomly selected unlabeled images. The generated pseudo labels were then checked by human annotators to validate the quality and high-quality ones are used to refine the CNN model. The process was repeated until the desired performance was achieved. The proposed approach was evaluated on a sorghum dataset, consisting of 1,269 manual-labeled images, yielding a coefficient of determination ($R^2$) of 0.8815 between the ground truth and predicted value. However, the random selection of unlabeled samples to be labeled may lead to sub-optimal and efficient solutions.

In \cite{siddique2022self}, a novel semi-supervised learning framework based on Panoptic FPN \citep{kirillov2019panoptic} was proposed for panoptic segmentation \citep{kirillov2019panoptic} of fruit flowers. To increase the sample diversity, a sliding window-based data augmentation approach \citep{dias2018multispecies} was employed to augment both the labeled and unlabeled samples. The model was firstly pre-trained on the COCO \citep{lin2014microsoft} and COCO stuff \citep{caesar2018coco} datasets and then fine-tuned on the labeled samples in a supervised manner. The trained model was then used to generate pseudo labels for the unlabeled samples and a robust segmentation refinement method (RGR) \citep{dias2018semantic} was adopted to refine the predicted score maps. The proposed method was evaluated on the multi-species flower dataset \citep{dias2018multispecies}, which contains four subsets: AppleA, AppleB, Peach, and Pear. The proposed algorithm with RGR refinement strategy yielded the highest IoU and F1 scores on the APPLEB, Peach, and Pear subsets and outperformed the supervised approach that was only trained with a small amount of labeled data. 

In \cite{zhang2022self}, self-distillation \citep{zhang2019your} with an advanced spectral-spatial vision transformer network \citep{dosovitskiy2020image} was proposed to accurately predict the nitrogen status of wheat with UAV images. The proposed spectral-spatial vision transformer contained a spectral attention block (SAB) and a spatial interaction block (SIB), which focused on the spectral and spatial information between the image patches, respectively, to fully capture the explicit encoding of patches. The teacher-student framework (Section~\ref{sec:semi}) was employed with the spectral-spatial vision transformer networks as the base models. The teacher model was updated by the student through the exponential moving average (EMA). Evaluated on a total of 1,449 field images of different growing stages collected by a UAV, the proposed approach achieved an overall accuracy of 96.2\% and outperformd the model trained without a semi-supervised training strategy with an accuracy of 94.4\%.

\textit{\textbf{Weakly-supervised learning. }}
The weakly-supervised learning algorithm, High-Performance Instance Segmentation with Box Annotations (Boxinst) \citep{tian2021boxinst}, was employed in \citep{qiang2022phenotype} for instance segmentation of leafy greens and phenotype tracking with only box-level annotations. To better distinguish leaf green vegetables from noisy backgrounds (e.g., shadow, dead grass, and soil), Excess Green (ExG) feature space was adopted instead of the LAB color space. Furthermore, post-processing methods, such as the area threshold method and k-means clustering \citep{hartigan1979algorithm} algorithm, were applied to filter weeds to achieve a clean instance segmentation of leafy greens. A multi-object tracking algorithm \citep{bewley2016simple} was then adopted to track the phenotypic changes of each vegetable for monitoring and analyzing purposes. Validated on a self-collected dataset, containing 656 training images (with box annotations) and 92 testing images (with pixel-level annotations), the proposed approach with ExG feature space representations yielded an F1-score of 95\% on instance segmentation and multi-object segmentation and tracking accuracy (MOTSA) of 91.6\%, outperforming the method with LAB with an F1-score of 92.0\% and MOTSA of 89.3\%.

To develop an autonomous harvester, \cite{kim2021weakly} proposed a weakly supervised crop area segmentation approach to identify the uncut crop area and its edge based on computer vision technologies with only image-level annotations. A four-layer CNN model followed by a global average pooling was trained to generate class activation maps (CAMs) \citep{zhou2016learning}, which were used for class-specific scoring (i.e., crop, harvested area, and backgrounds) by a Softmax layer. To evaluate the proposed framework, a self-collected crop/weed dataset, containing 1,440 training images and 120 testing images, was employed, showing that the proposed approach yielded the lowest inference time (less than $0.1~s$) and a comparable IoU value of 94.0\% -- outperforming the FCN \citep{long2015fully} algorithm with the inference time of $0.54~s$ and an IoU of 96.0\%. 
In \cite{adke2022supervised}, two supervised learning algorithms (Mask R-CNN \citep{he2017mask} and S-Count) and two weakly supervised approaches (WS-Count \citep{bellocchio2019weakly} (MIL-CAM based weakly supervised counting) and CountSeg \citep{cholakkal2020towards} (CAM \citep{zhou2016learning} based counting with partial labels)) were implemented and compared for cotton boll segmentation and counting using only image-level annotations.  Evaluation on a cotton dataset consisting of 4350 image patches showed that the weakly supervised approaches, WS-Count and CountSeg (RMSE values of 1.826 and 1.284, respectively), were able to achieve comparable performance as the supervised learning approaches, S-Count and Mask R-CNN \citep{he2017mask} (RMSE values of 1.181 and 1.175, respectively), while the weakly supervised approaches were at least $10\times$ cost efficient in labeling. 

In \cite{dandrifosse2022deep}, an automatic wheat ear counting and segmentation framework was developed based on the weakly-supervised learning algorithm \citep{birodkar2021surprising} to reduce labeling costs. Firstly, YOLOv5 \citep{jocher2020yolov5} model was trained on a self-collected dataset with bounding box annotations to localize the wheat ear. To avoid pixel-level annotations for the instance segmentation of wheat ear, Deep-MAC algorithm \citep{birodkar2021surprising} was employed to segment the ears in the obtained bounding boxes. Then, the ear counts were obtained by using the ear density map. The proposed approach yielded an average F1 score of 86.0\% for the wheat ear segmentation on a self-collected dataset.

In \citep{petti2022weakly}, multi-instance learning (MIL, Section~\ref{sec:wsl}) was employed for cotton blossom counting with aerial images. The images from an actual cotton field collected with a drone were divided into some small and overlapping patches (1/64 the size of an image) and partitioned into two subsets: subset A with point labels and subset B with binary image-level annotations indicating whether or not blossoms were present in the image patches. A binary classifier based on DenseNet-121 \citep{huang2017densely} was trained on the subset A with the MIL framework using the cross-scale consistency loss \citep{shen2018crowd} to indicate whether or not point annotations were present. To save expensive labeling time, the trained CNN model was used to generate annotations for the image patches in subset B, and these annotations were then verified by a human expert. Finally, both image patches in subsets A and B were used to fine-tune the model. The proposed approach achieved a minimum mean absolute count error (i.e., MAE) of 2.43, outperforming other CNN-based approaches such as VGG-16 \citep{simonyan2014very} (MAE of 2.90) and AlexNet \citep{krizhevsky2017imagenet} (MAE of 3.84). \vspace{10pt} \\
\underline{\textbf{No supervision}} \vspace{5pt} 

\textit{\textbf{Unsupervised representation learning. }}
In \cite{wang2018unsupervised}, a conditional probability distribution model, named conditional random field (CRF), based on the unsupervised hierarchical Bayesian model, Latent Dirichlet Allocation (LDA) model \citep{blei2003latent}, was proposed for plant organ (i.e., fruits, leaves, and stems) segmentation. The LDA was used to generate the initial segmentation labels by clustering pixels into different classes that are considered as the unary potential used in the CRF model. To improve the accuracy of image segmentation for different fruit growth stages, a multi-resolution CRF (MRCRF) algorithm was proposed to obtain multi-resolution features by down-sampling the images twice. Experimental results on a self-collected image showed that the proposed MRCRF was able to achieve high image segmentation accuracies. However, the approach was only evaluated on a small dataset with a few numbers of images (i.e., 9 images).
In \cite{zhang2018unsupervised}, LDA was also employed to segment leaves and greenhouse plants for plant phenotype analysis. The proposed approach was evaluated on the subset A1 of the CVPPP dataset \citep{scharr2014annotated}, achieving a high segmentation accuracy for segmenting greenhouse plants and leaves.

\begin{figure}[!ht]
  \centering
  \includegraphics[width=0.4\textwidth]{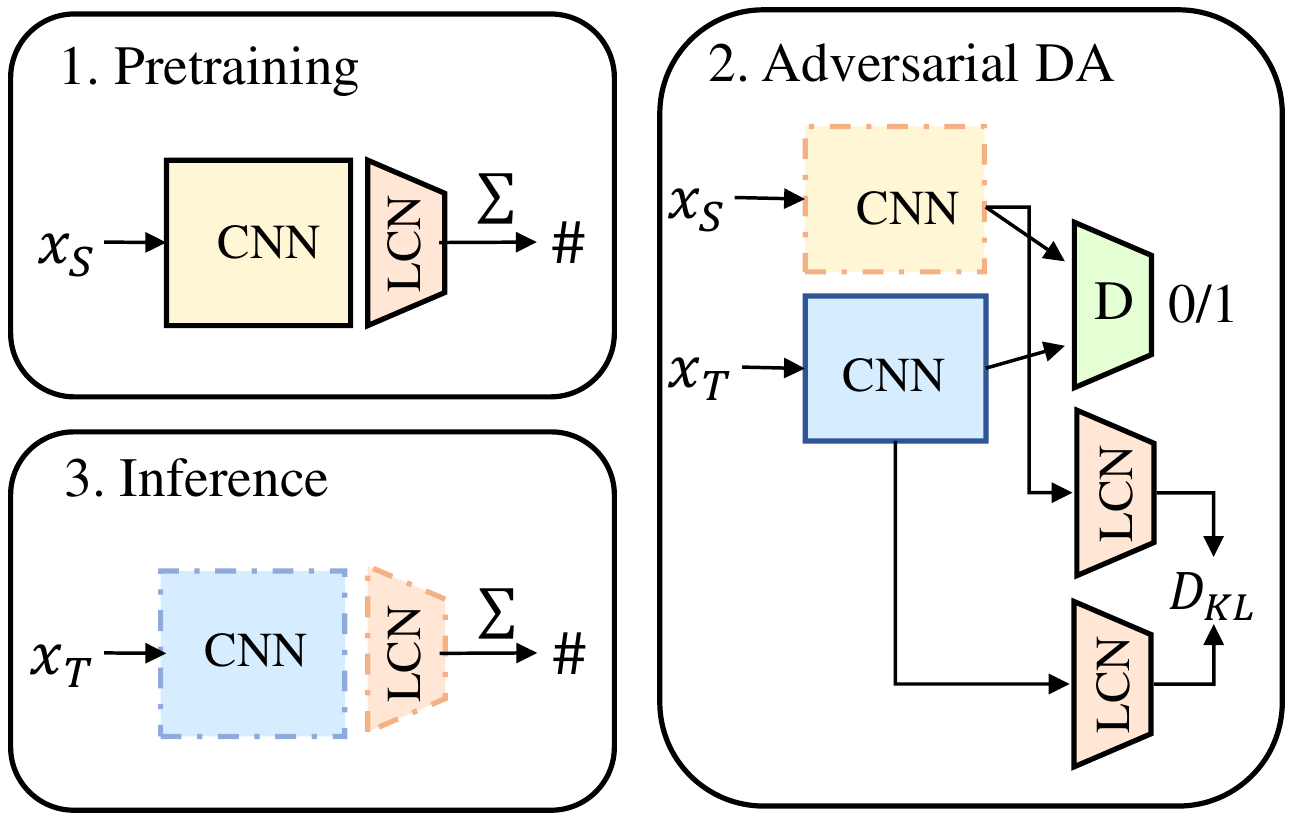}
  \caption{The unsupervised representation learning of domain shift \citep{giuffrida2019leaf}.}
  \label{fig:domain_shift}
  \vspace{-10pt}
\end{figure}

Domain shift refers to the difference between the source domains and target domains in the statistical distribution. Unsupervised learning is adapted to the problem of adapting a previously trained model on the source domain but testing on a new target domain without annotations (i.e., domain adaptation (DA)). In \cite{giuffrida2019leaf}, an unsupervised adversarial learning framework (ADDA) \cite{tzeng2017adversarial} was employed to reduce domain shift in leaf counting problem as shown in Fig.~\ref{fig:domain_shift}. First, the CNN model was pre-trained on the source domain in a supervised learning manner. Then, the feature representation of the source domain and target domain were fed into the adversarial learning network to minimize the domain shift and output the leaf counting using the adversarial loss. Lastly, leaf counting can be calculated on the target domain with the trained CNN model. The authors trained the model on the subset A1, A2, and A3 of the CVPPP dataset (source domain) \citep{scharr2014annotated} and tested on the MM dataset \citep{cruz2016multi} (Intra-species DA) and Komatsuna dataset \citep{uchiyama2017easy} (Inter-species DA), showing that the proposed approach significantly outperformed the baseline methods and obtained the lowest MSE of 2.36 and 1.84 for the intra-species and inter-species DA, respectively.
To address the laborious labeling and domain shift issues in plant organ counting for image-based plant phenotyping problems, an unsupervised domain-adversarial learning approach \citep{ganin2016domain} was employed in \citep{ayalew2020unsupervised}. The framework consisted of two parallel classification networks: one was designed for the source domain and another was designed to distinguish whether the samples come from the source domain or from the target domain. The proposed approach was evaluated on two domains: the wheat spikelet counting task (adapts from indoor images to outdoor images) \citep{alkhudaydi2022counting} and the leaf counting task (adapts from one plant species to another different plant species) \citep{giuffrida2019leaf}. Compared to the baseline model without domain adaptation, the proposed approach reduced the MAE and RMSE in the wheat spikelet counting experiment by 59.3\% and 58.0\%, respectively. Similarly, it yielded a 71.6\% drop of MSE in the leaf counting problem as compared to the baseline method.

Precision irrigation \citep{abioye2020review} aims to optimize the irrigation volume of each crop with minimum water volume. In \cite{tschand2023semi}, an intelligent irrigation system based on advanced computer vision algorithms was developed to analyze the crop color and optimize the irrigation volume for each crop. Image data acquired by a drone system was first passed to the k-means clustering \citep{hartigan1979algorithm} algorithm to isolate color clusters. Then, the clustered image features were used to train a recurrent neural network (RNN) to output predicted irrigation volume (PIV) for precision irrigation. The NASA PhenoCam Vegetation Phenology Imaging dataset \citep{seyednasrollah2019phenocam}, containing 393 site data, was adopted to validate the developed system, showing an average prediction accuracy of 89.1\% and an average ROC AUC of about 96.0\%, which was lower than the pre-defined error margins (5\%).


\textit{\textbf{Self-supervised learning. }} In \cite{lin2022self}, a novel self-supervised leaf segmentation framework was proposed without manually labeled annotations. Specifically, a lightweight CNN model was used to extract the pixel-level feature representations from the input images and output semantic labels. Then, the fully connected conditional random field (CRF) method \citep{krahenbuhl2011efficient} was adopted to refine the generated pseudo labels. After that, a color-based leaf segmentation algorithm was designed to identify leaf regions in the HSV color space. To rectify the distorted color in an image, a GAN-based pixel2pixel image translation network \citep{isola2017image} was employed for color correction, in which the generator learned to translate the input images with poor lighting conditions to the images with natural lighting conditions. The authors conducted the experiments on two open-sourced datasets, Plant Phenotype (CVPPP) and Leaf Segmentation Challenge (LSC) dataset \citep{minervini2016finely} and a self-collected dataset Cannabis dataset, showing that the proposed framework was able to achieve better or comparable performance than some mainstream unsupervised learning and supervised learning algorithms. For example, the proposed approach achieved a Foreground-Background Dice (FBD) score of 94.8\% on the Cannabis dataset, compared to the unsupervised approach EM \citep{kumar2012leafsnap} (FBD: 16.1\%) and supervised approach SYN \citep{ward2018deep} (FBD: 62.2\%).

\subsection{Postharvest quality assessment}
Machine vision and imaging technologies have become pervasive in the postharvest quality assessment of agricultural products, with applications ranging from automated grading and sorting based on shape, size, and color to more complex defect detection tasks \citep{CHEN2002173, Blasco2017, LU2020111318}. However, due to the significant biological variations in horticultural products, defect detection remains a challenging task that typically requires manual inspection. To overcome this limitation, researchers have turned to deep learning algorithms to enhance machine vision systems' capabilities. In particular, recent studies have explored the use of label-efficient learning algorithms (as summarized in Table~\ref{tab:table2}) for defect detection tasks, reducing the need for labeled training data and human efforts. 
\vspace{10pt} \\
\underline{\textbf{Weak supervision}} \vspace{5pt} \\
\textit{\textbf{Semi-supervised learning. }} In \cite{li2019greengage}, an ensemble stochastic configuration networks (SCNs, \cite{wang2017stochastic}) algorithm was employed for greengage grading with the semi-supervised co-training \citep{blum1998combining}. A self-collected dataset, consisting of 996 labeled images and 4,008 unlabeled images of four grades (excellent grade, a superior grade with scars, defective grade, and defective grade with scars), was used to validate the proposed approach. The unlabeled images were then pseudo-labeled by the semi-supervised co-training approach. SCNs were then trained on both labeled and pseudo-labeled samples with semantic error entropy measure constraints \citep{chen2017simulated}. SCNs achieved a recognition rate of 97.52\%, exceeding the accuracies obtained by CNN-based methods \citep{li2017intelligent} by 4\% and gaining 6\% improvement from traditional machine vision-based method \citep{jianmin2012spherical}.

\begin{figure}[!ht]
  \centering
  \includegraphics[width=0.48\textwidth]{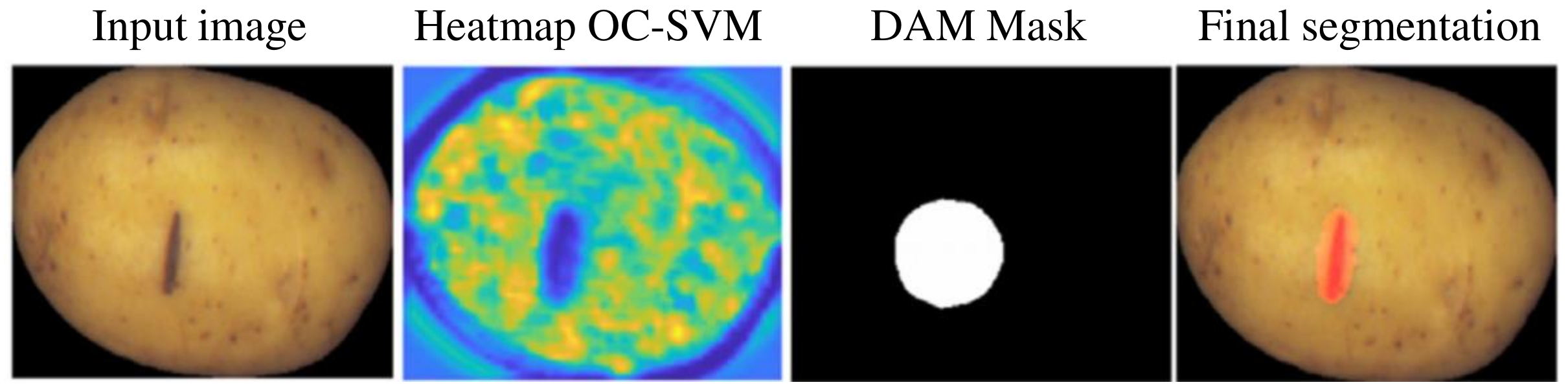}
  \caption{Example of the intermediate results of the weakly supervised approach for detecting and segmenting potato defects \citep{marino2019weakly}.}
  \label{fig:figure6}
  \vspace{-5pt}
\end{figure}

\begin{figure*}[!ht]
  \centering
\includegraphics[width=0.90\textwidth]{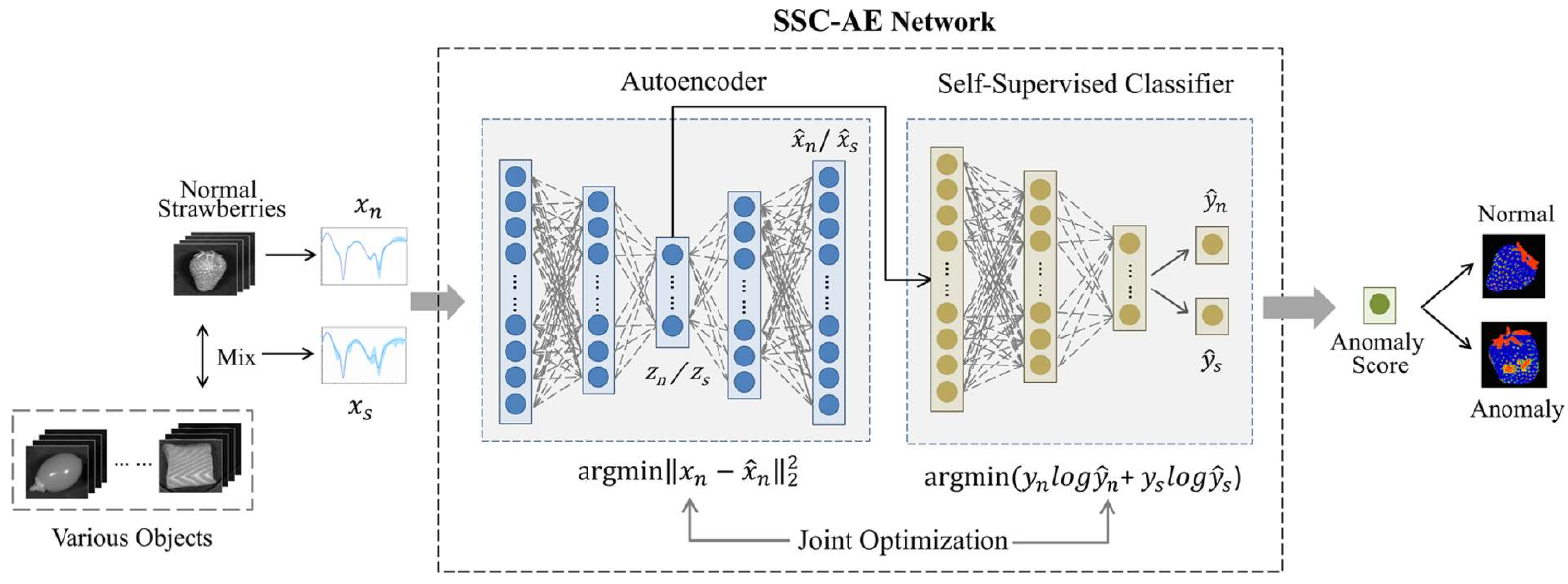}
  \caption{Architecture of the proposed self-supervised approach for anomaly
detection of strawberries \citep{liu2022joint}.}
  \label{fig:figure7}
  \vspace{-5pt}
\end{figure*}

\textit{\textbf{Weakly-supervised learning. }}
In \citep{marino2019weakly}, a weakly-supervised approach based on CAM \citep{zhou2016learning} (Section~\ref{sec:semi}) was proposed to detect and segment potato defects for meticulous quality control with only image-level annotations. To collect a face-wise dataset, each potato was captured with four images from different views and annotated by two experts into six categories: damaged, greening, black dot, common scab, black scurf, and healthy, resulting in 9,688 potato images. Based on gravity, damaged and greening potatoes were further classified into serious or light defects, finally resulting in a total of eight potato classes. Three CNN-based classification models (i.e., AlexNet \citep{krizhevsky2017imagenet}, VGG16 \citep{simonyan2014very}, and GoogleNet \citep{szegedy2015going}) were first trained on the face-wise dataset through transfer learning in a supervised manner. Inspired by CAM \citep{zhou2016learning}, the defect activation maps (DAMs) were extracted from the CNN models to classify the potential defects with only the image-level annotations. To further improve the segmentation of the defects, a one-class support vector machine (OC-SVM) was employed to identify the abnormal pixels within the DAMs. The intermediate results of the above process are shown in Fig.~\ref{fig:figure6}. Experimental results showed that the proposed approach achieved an F1-score of 94.0\%,  outperforming the conventional classifiers, e.g., the SVM classifier with an F1-score of 78.0\%. However, the datasets used in the study were collected in laboratory environments, which may not be feasible for practical usage with more complex lighting and operating environments in the real world. 
\vspace{10pt} \\
\underline{\textbf{No supervision}} \vspace{5pt} 

\begin{table*}[!ht]
    \centering
    \label{tab:table2}
    \caption{Application of label-efficient learning in plant phenotyping and post-harvest quality assessment.}
    \includegraphics[width=0.85\textwidth]{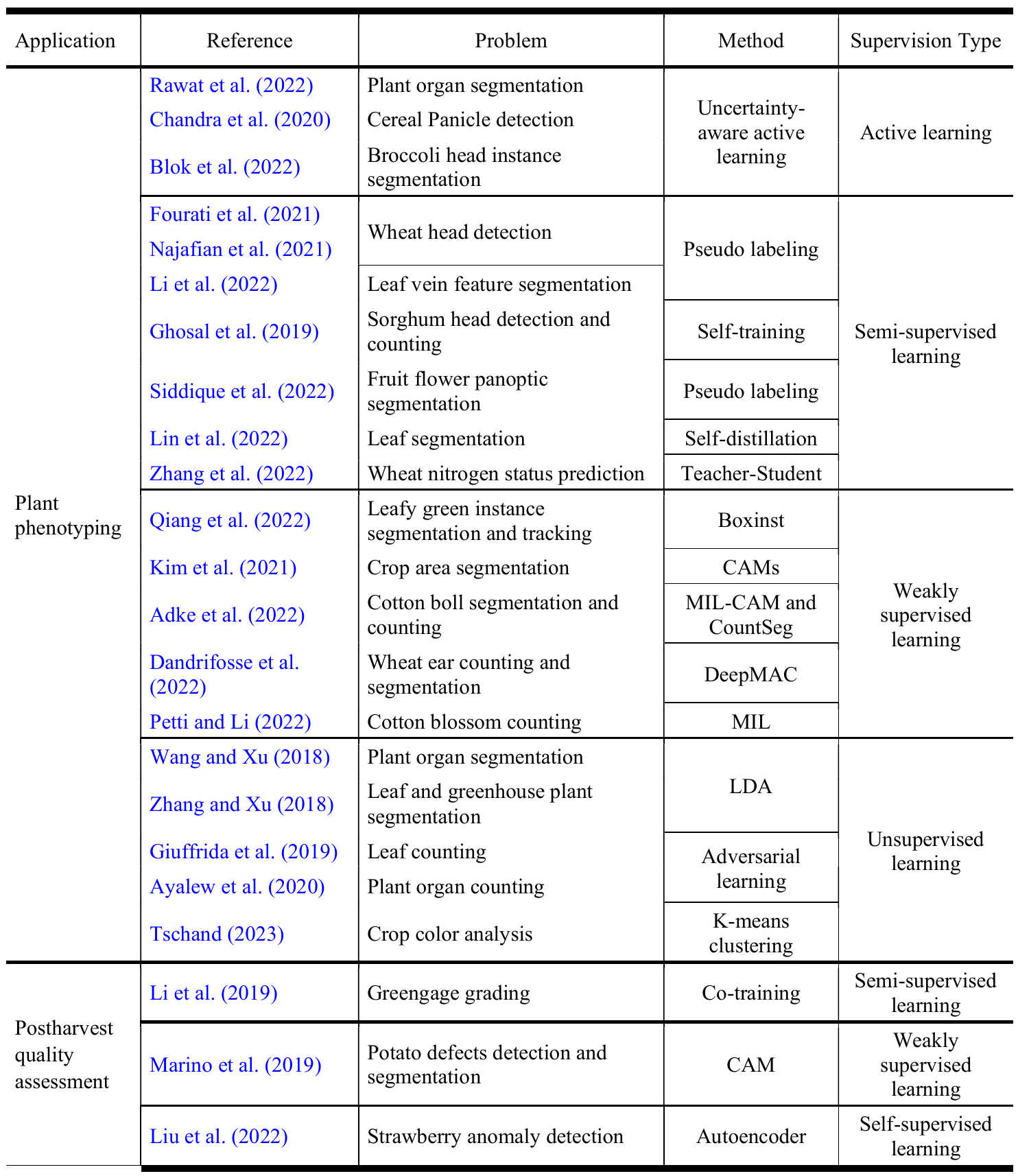}
\end{table*}

\textit{\textbf{Self-supervised learning. }}
In \cite{liu2022joint}, the authors proposed a novel self-supervised anomaly detection method, SSC-AE network, with hyperspectral strawberry data. A strawberry anomaly detection dataset, containing 601 normal and 339 anomalous strawberry samples, was collected with an NIR hyperspectral imaging instrument to validate the developed framework. As shown in Fig.~\ref{fig:figure7}, the SSC-AE network consisted of two components, an autoencoder (AE) and a self-supervised classifier (SSC), where the AE network was designed to extract the informative feature representations and the SSC network was trained to distinguish whether the learned feature representations come from the normal strawberries or synthetic anomalous strawberries generated by mixing the spectra of normal strawberries with various objects (e.g., milk powder, tomatoes, and grapes), which were updated with a joint optimization loss. Experimental results showed that SSC-AE achieved the highest anomaly detection performance with an AUC score of 0.908 ± 0.005 and an F1-score of 0.840 ± 0.005, topping the six baseline methods.


\section{Discussion and Future Research Directions}
\label{sec:dic}
Label-efficient learning has shown promising results in minimizing the need for annotated data and improving the accuracy of DL models in various applications, but there are still several unresolved issues related to training and evaluation that must be carefully considered to fully harness its benefits, such as pseudo-label refinement for unlabeled data, open-set learning from unlabeled data, continual learning from unlabeled data, and multi-modal learning from unlabeled data. 

\subsection{Pseudo-label refinement for unlabeled data}
Label-efficient learning algorithms utilize unlabeled data to facilitate training, and pseudo-labels are commonly employed for representation learning, such as semi-supervised learning and self-supervised learning, therefore it is crucial to keep up the quality of the pseudo-labels. Recent research focuses on removing unreliable samples within uncertain pseudo-labels \citep{sohn2020fixmatch, liu2021unbiased} to address the side effects of the unreliable pseudo-labels. For example, in Fixmatch \cite{sohn2020fixmatch}, a pseudo-label was  retained only if the model produces a high-confidence prediction. On the other hand, research like \citep{wang2021uncertainty, wang2022semi} focused on utilizing these unreliable and noisy pseudo-labels to enhance feature representation. For instance, a novel uncertainty-aware pseudo-label refinery framework was proposed in \cite{wang2021uncertainty} to progressively refine high-uncertainty predictions during the adversarial training process to generate more reliable target labels.

In future works, research could focus on developing novel algorithms that improve the quality of pseudo-labels, as well as examining the trade-offs between using reliable and unreliable pseudo-labels \citep{wang2021uncertainty}. Additionally, exploring the theoretical underpinnings of these approaches and the properties of pseudo-labels could provide insights into how to better design and use these methods \citep{sohn2020fixmatch}.

\subsection{Open-set learning from unlabeled data}
In label-efficient learning, the goal is to train a model with as few labeled samples as possible, while leveraging a large amount of unlabeled data. However, when dealing with open-set challenges, where the unlabeled data may contain unknown or unseen classes, the effectiveness of label-efficient learning may be greatly hindered \citep{chen2022semi, fontanel2022detecting}. Most existing label-efficient learning methods assume a closed-set scenario, where the unlabeled data comes from the same data distribution as the labeled data \citep{liu2022open}. However, in an open-set scenario, where the unlabeled data contains out-of-distribution (OOD) samples \citep{saito2017asymmetric, bousmalis2017unsupervised}, e.g., task-irrelevant or unknown samples, directly applying these label-efficient methods may lead to significant performance degradation due to catastrophic error propagation \citep{liu2022open}.

To address open-set challenges in label-efficient learning, recent works propose various sample-specific selection strategies to detect and then discount the importance or usage of OOD samples \citep{liu2022open, guo2020safe}. 
The pioneer works, such as UASD \citep{chen2020semi} and DS3L \citep{guo2020safe}, proposed dynamic weighting functions to down-weight the unsupervised regularization loss term proportional to the likelihood that an unlabeled sample belongs to an unseen class. Follow-up works, such as \cite{liu2022open}, added an additional OOD filtering process into the existing semi-supervised approaches during training to detect and discard potentially detrimental samples. For example, an offline OOD detection module based on DINO \citep{caron2021emerging} model was first pre-trained in a self-supervised way and fine-tuned with the available labeled samples. The OOD objects were then filtered out by computing the distance between the feature vectors of the image and the available labeled data. However, open-set label-efficient learning still faces many challenges, such as it is challenging to integrate OOD detection or novel class discovery with existing algorithms in a unified model to advance the selective exploitation of noisy unlabeled data \citep{liu2022open}.

In summary, open-set challenges in label-efficient learning require the development of new methods that can effectively handle OOD samples and unknown classes. Future research efforts should focus on developing unified models that can integrate OOD detection and novel class discovery with label-efficient learning and address the challenges posed by distribution mismatch \citep{saito2017asymmetric}, imbalanced class distribution \citep{chen2020semi}, and discovery of unseen classes  \citep{liu2022open} in real-world unlabeled data.

\subsection{Continual learning from unlabeled data}
Label-efficient learning with continual learning refers to the scenario where a model has to learn from limited labeled data and incrementally update its knowledge with new unlabeled data \citep{chen2022semi, wang2023comprehensive}. This is particularly relevant in real-world scenarios where data may be expensive to label or not readily available. In this context, continual learning (CL), also referred to as incremental learning or lifelong learning aims to extend the knowledge of an existing model without accessing previous training data \citep{chen2022semi, wang2023comprehensive}.

To prevent catastrophic forgetting when continuously updating the model, most CL approaches use regularization objectives to retain the knowledge of previous tasks \citep{mccloskey1989catastrophic}. However, in label-efficient learning scenarios, there is an additional challenge in not having access to all the unlabeled training data at once, due to, for instance, privacy concerns or computational constraints. One possible approach to label-efficient incremental learning is to use unlabeled data to estimate the importance weights of model parameters for old tasks, thus preventing catastrophic forgetting \citep{aljundi2018memory}. Another approach is to use knowledge distillation objectives to consolidate the knowledge learned from old data \citep{lee2019overcoming}. However, addressing challenges such as modeling new concepts and evolving data streams remains a nontrivial task. It also poses a new challenge to expand the representations for novel classes and unlabeled data. To this end, several strategies are adopted to dynamically update representations in the latent space, including creating new centroids by online clustering \citep{smith2019unsupervised} and updating the mixture of Gaussians \citep{rao2019continual}. Self-supervised techniques have also been applied to the unlabeled test data to overcome possible shifts in the data distribution \citep{sun2020test, wang2020tent}.

In summary, the open challenges in label-efficient learning with incremental learning include addressing catastrophic forgetting \citep{lee2019overcoming}, modeling new concepts \citep{chen2022semi}, and distribution shifts \citep{sun2020test}. Without access to all the unlabeled training data at once, directly applying many existing label-efficient learning methods may not guarantee good generalization performance. For instance, pseudo-labels may suffer from the confirmation bias problem \citep{arazo2020pseudo} when classifying unseen unlabeled data. Incremental learning from a stream of potentially non-i.i.d. unlabeled data also remains an open challenge in this area.

\subsection{Multi-modal learning from unlabeled data}
Multi-modal learning \citep{baltruvsaitis2018multimodal} from unlabeled data is a promising approach for improving model representation learning, where multiple modalities such as color, depth, and intensity are utilized to form discriminative (self-)supervision signals. For example, recent research has shown that the joint modeling of multi-modal data can be beneficial for agricultural applications, such as fruit detection \citep{gene2019multi}, weed recognition \citep{steininger2023cropandweed} and crop production enhancement \citep{sharma2022deepg2p}. Multi-modal learning combines multiple modalities, i.e., visual, audio, and text modalities, with label-efficient learning methods, such as semi-supervised \citep{cai2013heterogeneous}, self-supervised learning \citep{alayrac2020self}, and unsupervised learning \citep{hu2019deep} has been explored in general computer vision tasks. However, multi-modal learning from unlabeled data is still largely unexplored for agricultural applications. Future directions in this field involve addressing this semantic gap between modalities and developing more robust algorithms for multi-modal label-efficient learning in agricultural applications.


In conclusion, despite the significant progresses made in label-efficient learning in agriculture, there are still several challenges that remain, as discussed above. Nevertheless, if these challenges are adequately addressed and the opportunities presented by label-efficient learning are leveraged, it has tremendous potential to substantially reduce the cost and time required for data annotation. As a result, this could make deep learning models more accessible and practical for a diverse range of agricultural applications.

\section{Summary}
\label{sec:sum}
In recent years, the development of label-efficient methods has gained increased interest in agricultural research due to the high cost and difficulty of obtaining large-scale labeled datasets. This survey provided a principled taxonomy to organize these methods according to the degree of supervision, including methods under weak and no supervision. A systematic review of various applications in agriculture, such as precision agriculture, plant phenotyping, and postharvest quality assessment, was then presented. Through this survey, we highlighted the importance of label-efficient methods for improving the performance of ML/DL models with limited labeled data in agriculture and discussed the open challenges and future research directions in this area. The insights provided by this study can serve as a valuable resource for researchers and practitioners interested in developing label-efficient methods, ultimately leading to further advancements in the field. By providing an overview of recent developments and highlighting the potential of label-efficient methods in agriculture, this survey aims to stimulate further research in this important and exciting field.

\section*{Authorship Contribution}
\textbf{Jiajia Li}: Conceptualization, Investigation, Software, Writing – original draft;
\textbf{Dong Chen}: Conceptualization, Investigation, Software, Writing – original draft; 
\textbf{Xinda Qi}: Conceptualization, Investigation, Writing – original draft; 
\textbf{Zhaojian Li}: Supervision, Writing - review \& editing;
\textbf{Yanbo Huang}: Supervision, Writing - review \& editing; 
\textbf{Daniel Morris}: Writing - \& review \& editing. 
\textbf{Xiaobo Tan}: Writing - \& review \& editing. 


\typeout{}
\bibliography{ref}
\end{document}